\documentclass[10pt,twocolumn,letterpaper]{article}

\usepackage{cvpr}
\newcommand\mypara[1]{\noindent\textbf{#1.}}

\usepackage{soul}           

\newcommand{\ourmethod}{HorizonWeaver\xspace}

\newcommand{\ourdataset}{WeaverScape}

\usepackage{listings}          

\usepackage{dblfloatfix}
\usepackage{float}
\usepackage{multirow}

\definecolor{cvprblue}{rgb}{0.21,0.49,0.74}
\usepackage[pagebackref,breaklinks,colorlinks,allcolors=cvprblue]{hyperref}

\def\paperID{}
\def\confName{CVPR}
\def\confYear{2025}

\title{HorizonWeaver: Generalizable Multi-Level Semantic Editing for Driving Scenes}

\author{
Mauricio Soroco\\
Simon Fraser University\\
\and
Francesco Pittaluga\\
NEC Labs America\\
\and
Zaid Tasneem\\
NEC Labs America\\
\and
Abhishek Aich\\
NEC Labs America\\
\and
Bingbing Zhuang\\
NEC Labs America\\
\and
Wuyang Chen\\
Simon Fraser University\\
\and
Manmohan Chandraker\\
NEC Labs America\\
\and
Ziyu Jiang\\
NEC Labs America\\
}

\begin{document}
\maketitle
\begin{abstract}

Ensuring safety in autonomous driving requires scalable generation of realistic, controllable driving scenes beyond what real-world testing provides. Yet existing instruction-guided image editors, trained on object-centric or artistic data, struggle with dense, safety-critical driving layouts.

We propose \ourmethod, which tackles three fundamental challenges in driving scene editing: (1) \textbf{multi-level granularity}, requiring coherent object- and scene-level edits in dense environments; (2) \textbf{rich high-level semantics}, preserving diverse objects while following detailed instructions; and (3) \textbf{ubiquitous domain shifts}, handling changes in climate, layout, and traffic across unseen environments.
The core of \ourmethod\ is a set of complementary contributions across data, model, and training: \textbf{(1) Data: Large-scale dataset generation}, where we build a paired real/synthetic dataset from Boreas~\cite{burnett_ijrr23}, nuScenes~\cite{caesar2020nuscenes}, and Argoverse2~\cite{wilson2023argoverse2} to improve generalization; \textbf{(2) Model: LangMasks for fine-grained editing}, where semantics-enriched masks and prompts enable precise, language-guided edits; and \textbf{(3) Training: Content preservation and instruction alignment}, where joint losses enforce scene consistency and instruction fidelity.
Together, \ourmethod provides a scalable framework for photorealistic, instruction-driven editing of complex driving scenes, collecting 255K images across 13 editing categories and outperforming prior methods in L1, CLIP, and DINO metrics, achieving +46.4\% user preference and improving BEV segmentation IoU by +33\%. Project website: \href{https://msoroco.github.io/horizonweaver/}{https://msoroco.github.io/horizonweaver/}

\end{abstract}    
\section{Introduction}
\label{sec:intro}

\begin{figure*}[t]
\vspace{-2em}
\centering
\includegraphics[clip, trim=0cm 0cm 0cm 0cm, width=0.7\textwidth]{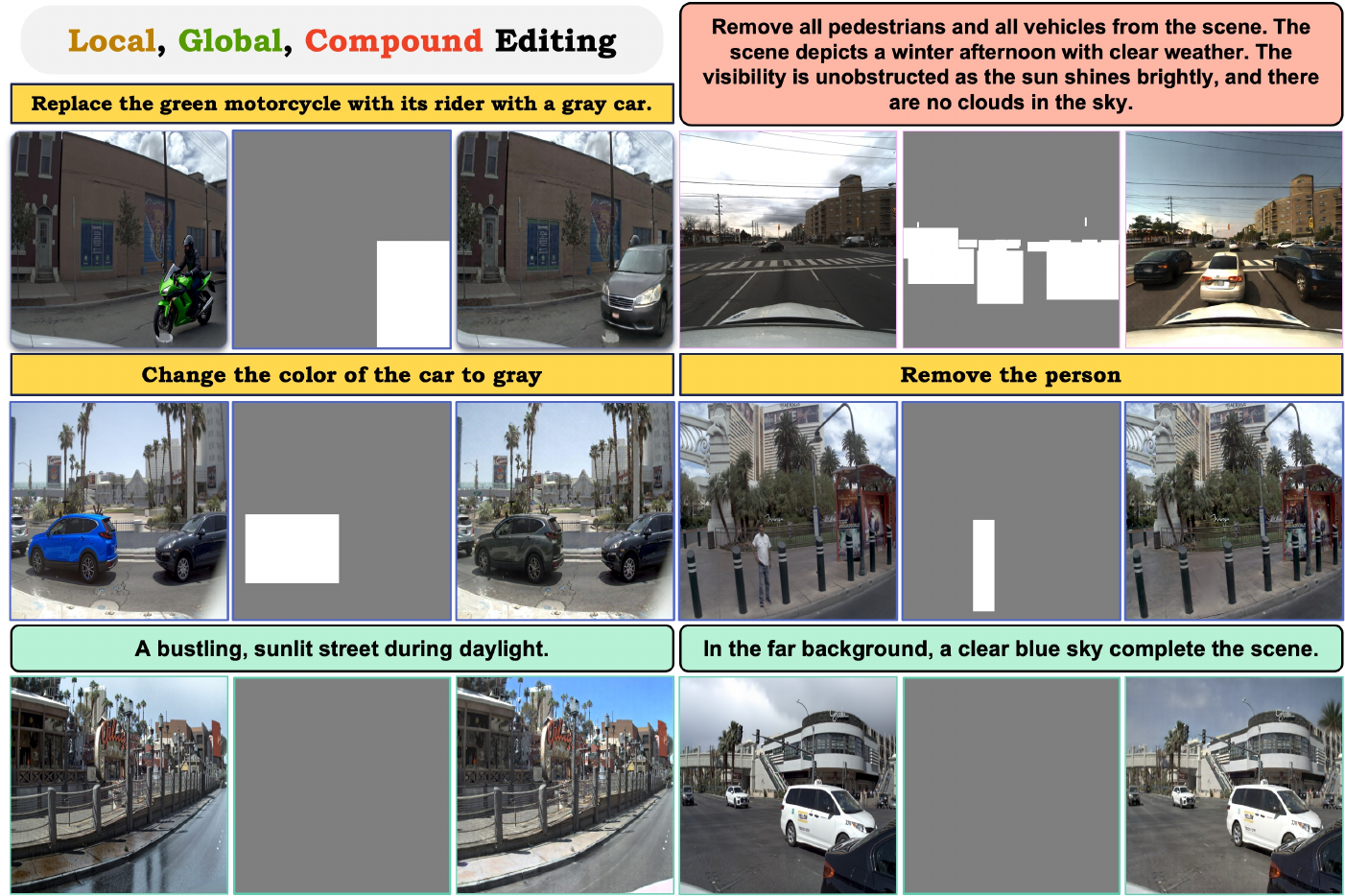}
\vspace{-0.5em}
\captionsetup{font=small}
\caption{In each example, from left to right are the input image, LangMasks, and output image. Masks for global edits are blank.
The careful design of \ourmethod across data (Section~\ref{sec:automatic-paired-dataset-generation}), model (Section~\ref{sec:LangMasks for fine-grained editing}), and training (Section~\ref{sec:methods-training-objectives}) addresses three critical challenges in driving scene editing: 1) mult-level granularity (rows 2, 3); 2) rich high-level semantics (rows 1, 2); 3) ubiquitous domains shifts (Tab.~\ref{tab:combined_OOD_editing_performance}).
}
\label{Fig:teaser}
\vspace{-1em}
\end{figure*}

Ensuring safety is a central challenge in deploying autonomous driving. Real-world testing within the Operational Design Domain (ODD) is limited by dynamic factors such as lighting, weather, traffic, and road conditions, making comprehensive data collection infeasible~\citep{mehlhorn2023ruling}. Generative models offer a scalable alternative for synthesizing diverse environments~\citep{gao2023magicdrive}, and instruction-guided image editing in particular enables fine-grained, language-based control while preserving realism through large-scale pretrained sources~\citep{ramesh2022dalle2,dalle3,Rombach2022StableDiffusion,brooks2023instructpix2pixlearningfollowimage,zhao2024ultraeditinstructionbasedfinegrainedimage}.

In traditional image editing domains, most image editing frameworks focus either on sparse, object-centric modifications~\citep{brooks2023instructpix2pix,zhao2024ultraeditinstructionbasedfinegrainedimage} or on global scene-level transformations treated in isolation.
However, \emph{driving scenes differ fundamentally} in the following three aspects:

\mypara{1. Multi-Level Granularity}
Autonomous driving scenes
are dense, containing numerous interacting entities such as vehicles, pedestrians, and traffic infrastructure, together within structured environments.
Effective editing thus requires \textbf{simultaneous precision at both local (object-level) and global (scene-level) scales} (e.g. adjusting traffic flow and changing weather while keeping the road layout intact).

\mypara{2. Rich High-Level Semantics}
Existing instruction-guided image editing struggles with two properties critical for autonomous driving scenes: \textbf{Content Preservation} and \textbf{Instruction Alignment}. \textbf{Content preservation} requires retaining unedited scene elements (e.g., lanes, signs, vehicles), avoiding harmful distortions to safety-critical cues. \textbf{Instruction alignment} demands accurately following detailed, multi-attribute language instructions, enabling controlled, combinatorial exploration of diverse driving conditions.

\mypara{3. Ubiquitous Domain Shifts}
Real-world driving scenes vary widely in climate, urban layout, traffic composition, and infrastructure, yet existing datasets cover only narrow geographic and environmental slices. Datasets such as nuScenes~\cite{caesar2020nuscenes}, Argoverse2~\cite{wilson2023argoverse2}, Waymo~\cite{Sun_2020_CVPR}, and Boreas~\cite{burnett_ijrr23} concentrate on a few cities or weather regimes, inducing domain bias and hindering generalization of editing models to unseen environments.

These bottlenecks motivate our key research questions:
\begin{center}
\fbox{
    \parbox{0.86\linewidth}{
        \textit{\textbf{Q1}: How to support global and local editing operations on driving scenes at multi-level granularity?}
        \newline
        \textit{\textbf{Q2}: How to guide editing models to faithfully follow instructions with rich, high-level semantics?}
        \newline
        \textit{\textbf{Q3}: How to achieve generalizable image editing across driving environments under domain shifts?}
    }
}
\label{core_questions}
\end{center}

To address these questions, we propose \textbf{HorizonWeaver} (Fig.~\ref{Fig:teaser}), a multi-granularity, semantically rich driving scene editing pipeline that generalizes to unseen domains and introduces novel designs on all three fronts: \emph{data, model, and training}.
\textbf{1) Data: Automatic Paired Dataset Generation.}
We build an instruction-guided editing dataset with paired real/synthetic driving images from Boreas~\cite{burnett_ijrr23}, nuScenes~\cite{caesar2020nuscenes}, and Argoverse2~\cite{wilson2023argoverse2}, enhancing generalization across shifting, unseen driving domains.
\textbf{2) Model: LangMasks for Fine-Grained Editing.}
We introduce LangMasks, a pixel-level conditioning mechanism for localized, language-guided edits, enabling instruction-aligned, content-preserving manipulation of specific objects (insert, remove, replace, modify) in densely populated driving scenes.
\textbf{3) Training: Content Preservation and Instruction Alignment.}
We design losses that jointly enforce scene consistency and instruction fidelity, enhancing perceptual editing quality and improving performance on safety-critical downstream tasks such as BEV (Bird's-Eye View) map segmentation.
We summarize our concrete contributions below
\begin{enumerate}[leftmargin=*]
\setlength\itemsep{0em}
\item \textbf{Comprehensive Datasets for Driving Scene Editing.}

In total, we collect 255K images covering 13 categories of driving-scene objects.

\item \textbf{Superior Driving Scene Editing Performance.}
HorizonWeaver consistently outperforms recent image editing models on unseen driving scenarios, achieving up to -45.5\% L1 distance, +12.1\% CLIP similarity, and +24.1\% DINO similarity improvements.

\item \textbf{Consistent User Preference.}
Driving scene images edited by HorizonWeaver are consistently preferred by real users (via voting), outperforming the second-best method by 46.4 percentage points.

\item \textbf{Benefits to Downstream Driving Tasks.}
Including driving scene images edited by HorizonWeaver further boosts downstream BEV map segmentation performance by 33\% IoU.

\end{enumerate}

\section{Related Work} \label{sec:related_works}

\mypara{Image Editing Dataset}
Developing image-editing datasets is more challenging than Text-2-Image~\citep{dalle3,Ramesh2021DALLE,ramesh2022dalle2,Rombach2022StableDiffusion}, with data scarcity being a major bottleneck~\citep{wang2023editbench,hui2024hq}. MagicBrush~\citep{zhang2024magicbrush} uses manual annotation with DALL-E2~\citep{ramesh2022dalle2}, while InstructPix2Pix~\citep{brooks2023instructpix2pix} generates pairs using the prompt-to-prompt method~\citep{Hertz2022Prompt2prompt} on LAION-Aesthetics~\citep{schuhmann2022laion5b}. SeedEdit~\citep{shi2024seededit} iteratively refines data and models. 
UltraEdit~\citep{zhao2024ultraeditinstructionbasedfinegrainedimage} automatically generates freeform and masked image editing pairs using LLM augmented image captions. 
Recently, No-Pair-edit~\citep{kumari2025learningimageeditingmodel} enables training image editors without paired images by using VLM and distribution matching supervision.
While this approach reduces dependence on paired datasets, it also limits its ability to maintain fidelity in unedited regions.
While prior work has made progress in localized object edits, these methods remain insufficient for the multi-object, context-rich street scenes.
\ourdataset, an instruction-driven editing dataset,
addresses this gap with instruction-guided edits that scene consistency and semantic alignment.

\mypara{Image Editing via Generation} 
Instruction-based editing of real photos is a key task in image processing~\citep{CLOVA, Crowson2022VQGAN-CLIP, Liu2020Open-Edit, zhang2023controlnet, Ruiz2022DreamBooth, pan2024kosmosg}. Large-scale diffusion models have greatly enhanced text-driven editing~\citep{Kawar2022Imagic, Saharia2022Imagen, li2023blipdiffusion, chen2023suti, ma2023subjectdiffusion, Meng2022SDEdit, mokady2023null, tumanyan2022plugandplay, Nichol2022GLIDE, sheynin2023emu}. Recent models like InstructPix2Pix~\citep{brooks2023instructpix2pix} and HIVE~\citep{zhang2023hive} allow users to edit images via instructions.  MagicBrush~\citep{zhang2024magicbrush} enhances this with manual annotations.
BAGEL~\citep{deng2025emergingpropertiesunifiedmultimodal} is a recent work that leverages multimodal data beyong text-image pairs to perform image editing.
Qwen-Image~\cite{wu2025qwenimagetechnicalreport}, trained on a combination of text-to-image, text-image-to-image, and image-to-image tasks, sets a new benchmark for maintaining the visual fidelity of unedited regions while enabling meaningful semantic edits. Unlike traditional image editors that either focus on sparse, object-centric, or global-level edits in isolation, our work is tailored for dense driving scenes.
As shown in Fig.~\ref{Fig:driving-features-are-dense}, driving scenes are densely populated, making it difficult for natural language alone to unambiguously describe fine-grained changes. In contrast, LangMasks embed text instructions within localized regions, providing semantic guidance for vehicles, pedestrians, and traffic elements while constraining edits to relevant areas.
We also contribute novel generative modeling techniques—specifically, an unsupervised loss that extends cycle-consistency to instruction-aligned editing, unlike \cite{Zhu2017CycleGAN} which is limited to fixed domains.
We show our dataset further boosts these methods in street-scene editing.

\mypara{Image Editing for Autonomous Driving} 
Rising demand for driving-scene data has led to 3D/4D generative and reconstruction-based editing methods ~\citep{liang2025driveeditor,yang2023unisim,sun2024lidarf,tonderski2024neurad,chen2024omnire, li2025unisceneunifiedoccupancycentricdriving, schneider2025neuralatlasgraphsdynamic, yang2025genassets}, though these struggle with diverse scene composition. Meanwhile, multi-condition generation methods are gaining interest~\citep{swerdlow2024BEVGen,yang2023bevcontrol,wang2023drivedreamer,gao2023magicdrive,wen2024panacea,alhaija2025cosmos,gao2024magicdrivedit,lu2024infinicube}. However unlike these works which train specialized condition to image models, we focus on a unified multi-task solution for flexible content preservation in instruction-guided image editing.
Our focus is a lightweight, camera-level editing pipeline that can be readily integrated into existing perception workflows for autonomous system evaluation. 

\section{\ourmethod}

\subsection{Overview}

\begin{figure*}[t]
  \centering
  \vspace{-1em}
  \captionsetup{font=small}
  \includegraphics[clip, trim=0cm 0.2cm 0cm 0.5cm, width=0.95\textwidth]{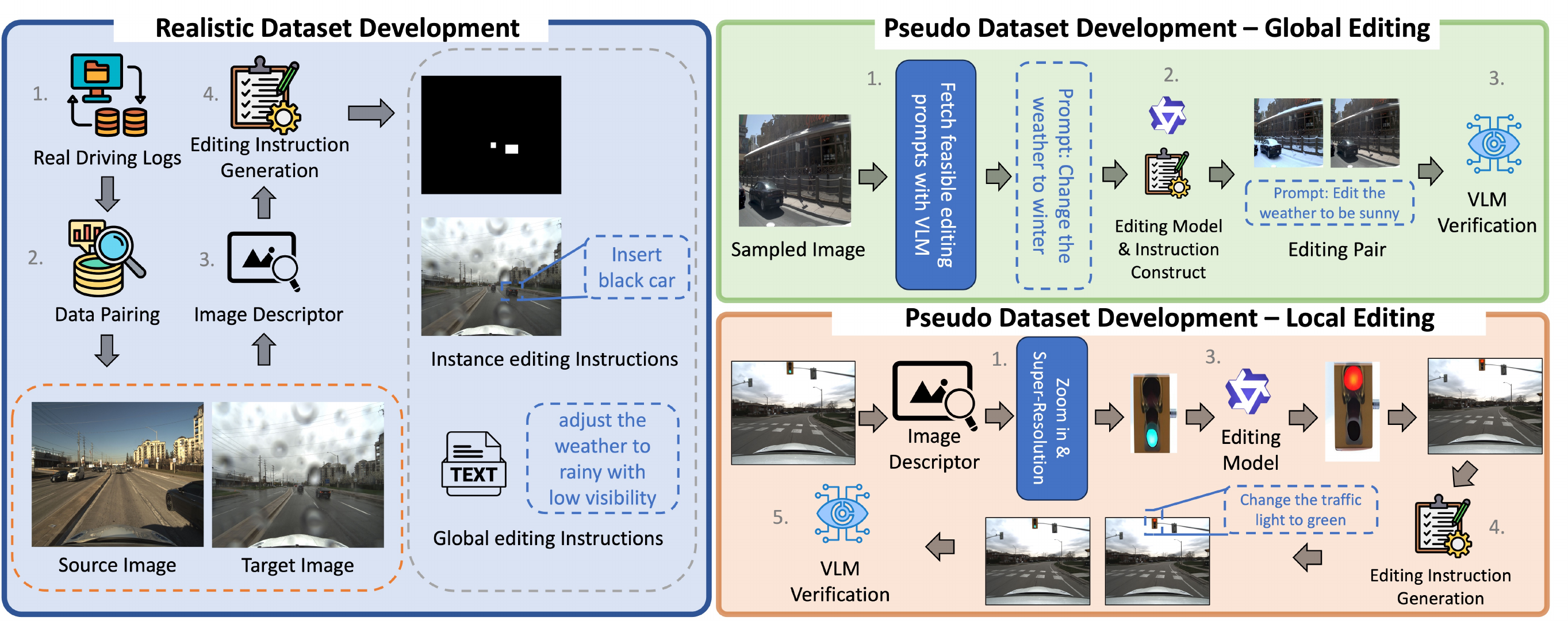}
  \vspace{-0.5em}
  \caption{\textbf{Dataset Construction}.
  {\setlength{\fboxsep}{1pt}\colorbox{blue!13}{Real-world data}} (Sec.~\ref{subsec:real-world-paird-data}) are paired by camera pose and annotated using an image descriptor pipeline and passed to an LLM to produce instructions.
  Pseudo-data (Sec~\ref{sec:methods-pseudo-dataset-development}) for {\setlength{\fboxsep}{1pt}\colorbox{orange!8}{Local edits}}  crop an annotated object before VLM filtering; {\setlength{\fboxsep}{1pt}\colorbox{green!8}{global edits}}  apply VLM filtering to full images.
  Our dataset is composed of image pairs, global editing instructions, and masks indicating fine-grained edits to perform. 
  }
  \label{Fig:pipeline}
  \vspace{-2em}
\end{figure*}

To cover our design on \textit{data}, \textit{model} and \textit{training}, we begin by introducing our curated dataset in Section~\ref{sec:automatic-paired-dataset-generation}, followed by our model design featuring an innovative LangMask in Section~\ref{sec:LangMasks for fine-grained editing}. Finally, we present our training strategy, which combines supervised and unsupervised objectives, in Section~\ref{sec:methods-training-objectives}.

\subsection{Diverse Paired Dataset for Driving Scene}\label{sec:automatic-paired-dataset-generation}

The construction of the dataset involves real-world image pairs and real-synthetic pseudo image pairs. For both, we use a detailed annotation process to capture varying edits in driving scenes while maintaining consistency. We overview our dataset construction pipeline in Fig.~\ref{Fig:pipeline}.
In this section, we explain our data annotation process and how we construct pixel masks for fine-grained editing.

\subsubsection{Real-World Paired Data}\label{subsec:real-world-paird-data}

Our first data contribution is a large-scale collection of paired real-world driving images enriched with hierarchical, fine-grained annotations. As illustrated in the ``Realistic Dataset Development'' section of Fig.~\ref{Fig:pipeline}, we construct pose-aligned image pairs from multi-season autonomous driving logs.

\begin{enumerate}

\item \textbf{Real Driving Logs.} We begin with multi-season autonomous driving logs featuring repeated routes and calibrated camera poses (e.g., Boreas~\cite{burnett_ijrr23}).

\item \textbf{Data Pairing.} From these originally unpaired recordings, we construct pose-aligned image pairs by searching, for each source frame, a target frame captured from nearly the same camera pose but during a different traversal of the route. A formal definition of the pairing procedure is provided in Appendix~\ref{app_sec:real-world-data-pairing-details}. This alignment ensures that the differences between two images primarily stem from environmental factors—such as weather, lighting, and traffic—rather than viewpoint changes, which is essential for learning realistic, controllable edits across diverse conditions including rain, snow, and fog.

\item \textbf{Image Descriptor.} A key challenge arises from substantial traffic variations across timestamps, making it difficult to generate precise and fine-grained scene-editing descriptions (see Fig.~\ref{Fig:driving-features-are-dense}). 
To address this, we introduce a training-free \emph{Image Descriptor} pipeline, an image-based analogue of iFinder~\cite{yao2025ifinder}. This pipeline integrates several pretrained models—such as object detectors~\citep{minderer2024scalingopenvocabularyobjectdetection}, segmentation models~\citep{kirillov2023segment-anything}, and Vision-Language Models (VLMs)~\citep{chen2024internvl}—to extract both global attributes (weather, time of day, scene type) and dense instance-level semantics.
These structured attributes enable the generation of rich annotations for global and local editing tasks. (Further implementation details are provided in the Appendix.)

\item \textbf{Editing Instruction Generation.} Using the outputs of the Image Descriptor, we construct global editing instructions by using ChatGPT-4o-mini to summarize the differences in global attributes between the source and target images, producing instructions such as ``adjust the weather to rainy with low visibility.''
Additionally, we instruct the model to remove all pedestrians and vehicles, as these elements inherently differ across timestamps.
For each instance appearing in the target image, we generate fine-grained insertion prompts based on its local attributes, such as ``insert a black car'' (see Fig.~\ref{Fig:pipeline}).

\end{enumerate}

This structured, hierarchical annotation pipeline provides high-quality instruction for both global scene adjustments and fine-grained instance manipulations.

\subsubsection{Pseudo-Paired Data for Diverse Scenes}\label{sec:methods-pseudo-dataset-development}

Real-world driving scenes vary widely across geography, weather, scene composition, and object appearances, posing a challenge for models trained on limited paired data. To address this, we extend our dataset construction pipeline to include additional variation in geographic and environmental attributes by incorporating unpaired datasets. This extension also enables the creation of controlled samples in which all or only part of the traffic is held fixed.

In the following, we introduce two data curation pipelines: global editing and local editing.

\mypara{Global Editing}
We construct pseudo global editing pairs using modern Vision--Language Models (VLMs)~\cite{chen2024internvl} and image-editing models. As shown in the Global Editing module of Fig.~\ref{Fig:pipeline}, the process consists of three steps:

\begin{enumerate}
\item \textbf{Prompt Generation.}
For each image sampled from driving logs, we use a VLM to extract global attributes—weather, time of day, and season—by selecting the closest match from a predefined set.
We then randomly sample a different attribute to form an editing instruction (e.g., converting a \emph{sunny, daytime, summer} scene to \emph{“snowy”}).

\item \textbf{Editing and Instruction Construction.}
An editing model~\cite{wu2025qwenimagetechnicalreport} applies the generated instruction to synthesize an edited image. Given these results may lack perfect photorealism, we treat the edited image as the source and the original real image as the target during training, encouraging the model to learn realistic appearance. Final natural-language instructions are obtained using VLM-based captions of the sampled image.

\item \textbf{VLM-Based Verification.}
To ensure correctness, a VLM verifies that the edited and original images correspond to the same underlying scene and that the intended global appearance change is successfully applied.
\end{enumerate}

\mypara{Fine-Grained Editing}
In fine-grained editing, we start from annotating each image with global scene attributes and instance-level object descriptions using our ``image descriptor'' (Section~\ref{subsec:real-world-paird-data}). For each object category of interest—vehicles, pedestrians, and traffic lights—we generate fine-grained edits through the following steps as shown in ``Local Editing'' part of Fig.~\ref{Fig:pipeline}:

\begin{enumerate}
    \item \textbf{Object Selection and Cropping.}
    We choose an annotated object, extract its 2D bounding-box crop, and enhance the crop using the InvSR super-resolution model~\cite{yue2025arbitrarystepsimagesuperresolutiondiffusion}.

    \item \textbf{Action and Prompt Generation.}
    We sample an editing action from \{insert, delete, modify, replace\}. A target object description is then selected or sampled, and an editing prompt is constructed (details in Appendix~\ref{App:psuedo-image-pipeline}).

    \item \textbf{Image Editing and Blending.}
    The enhanced crop is edited using Qwen-Image-Edit~\cite{wu2025qwenimagetechnicalreport}, resized, and Poisson-blended~\cite{perez2003poissonblending} back into the original image to form a pseudo-edited sample.

    \item \textbf{Editing Instruction Generation.}
    We generate instance-level editing instructions following the protocol in Section~\ref{subsec:real-world-paird-data}.

    \item \textbf{Quality Assurance.}
    To ensure reliable supervision, we apply an automated quality-control pipeline combining VLM-based semantic checks with low-level similarity metrics. The VLM verifies edit plausibility (e.g., realistic appearance, correct orientation, successful insertion or removal), while structural checks filter out artifacts across vehicles, pedestrians, traffic lights, and generic removals. This yields a clean, high-quality pseudo-edited dataset for training and evaluation (Appendix~\ref{app_sec:quality-control-local-edit}).
\end{enumerate}

\subsection{Fine-Grained and Instructive LangMasks}\label{sec:LangMasks for fine-grained editing}

Unlike common image-editing scenarios that are predominantly object-centric, editing autonomous driving scenes requires multi-level granularity, especially for realistic manipulations in dense environments (Fig.~\ref{Fig:driving-features-are-dense}). Traditional instruction formats such as natural language alone or binary masks~\cite{zhao2024ultraeditinstructionbasedfinegrainedimage} often fail to specify edits with sufficient precision. To this end, we propose LangMasks, a pixel-level conditioning mechanism that facilitates precise, language-guided, and content-preserving manipulation of individual objects even in densely populated driving scenes.

\begin{figure}[H]
  \centering
  \vspace{-0.4cm}
  \captionsetup{font=small}
  \includegraphics[clip, trim=0.1cm 0cm 0cm 0cm, width=0.47\textwidth]{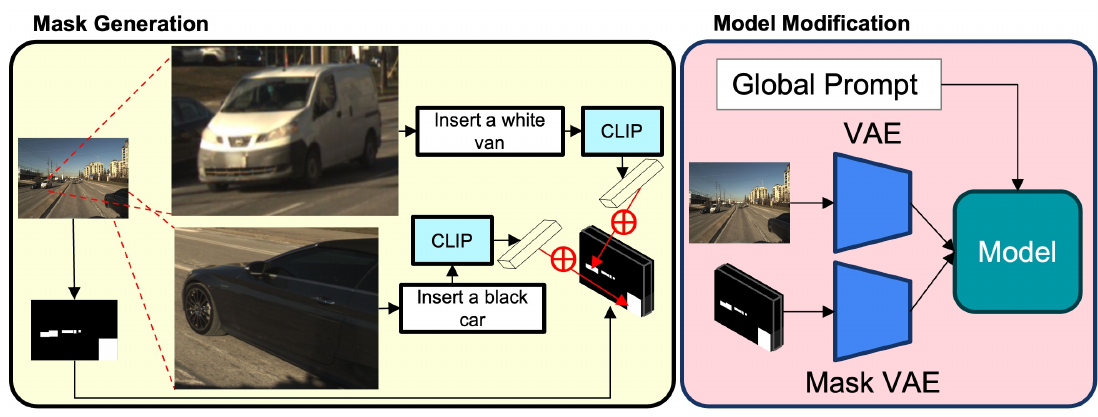}
  \vspace{-0.3cm}
  \captionsetup{font=small}
  \caption{\textbf{LangMask Generation and Training}.
    Left: To provide fine-grained instructions with rich semantics, we insert CLIP text features into binary masks (Section~\ref{sec:LangMasks for fine-grained editing}). 
    Right: To support LangMasks, we copy and expand the VAE. It is trained end to end with the editing model.}
  \label{Fig:model_mask_generation}
  \vspace{-1em}
\end{figure}

As shown in Fig.~\ref{Fig:model_mask_generation}, given the instance-level instructions in our data pairs, we convert each object-specific edit command (e.g., “Insert a white van”) into a CLIP embedding and insert it along the channel dimension for all pixels within the object’s bounding box, forming its LangMask. This representation precisely identifies the target object for editing while strictly preserving non-target objects and background content. Further details and discussion are provided in Appendix~\ref{app_sec:langMasks}.

\mypara{Model Architecture with LangMasks} 
To effectively incorporate LangMasks into the diffusion model, we repurpose the pretrained VAE encoder to process them directly. Specifically, we extend the VAE’s input layer with additional channels to accommodate the LangMask representation. During training, this augmented VAE encoder is jointly optimized with the diffusion model, while the original VAE weights remain frozen.

\subsection{Content Preservation \& Instruction Alignment}\label{sec:methods-training-objectives}

\begin{figure}[!b]
  \centering
  \vspace{-0.4cm}
  \includegraphics[clip, trim=0cm 5.5cm 0cm 0cm, width=0.475\textwidth]{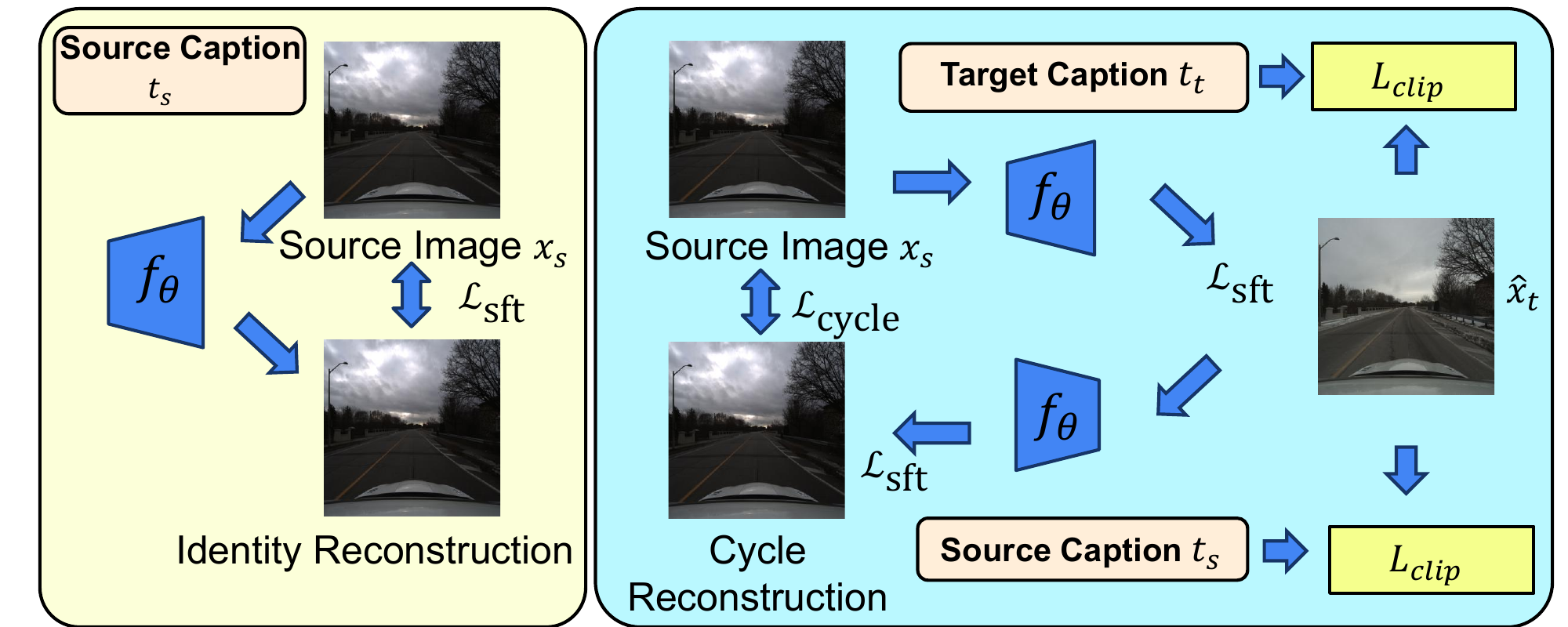}
  \vspace{-0.6cm}
  \captionsetup{font=small}
  \caption{\textbf{Training language-guided driving scene image editing}. Our training pipeline supports both supervised training for paired images and unsupervised training for unpaired ones (e.g. downstream unseen real scenarios). We include three training objectives: supervised fine-tuning $\mathcal{L}_\text{sft}$ (Section~\ref{sec:sft}), cycle consistency $\mathcal{L}_\text{cycle}$ (Section~\ref{sec:cycle_consistency}), and $\mathcal{L}_\text{clip}$ (Section~\ref{sec:clip_loss}).}
  \label{Fig:training_pipeline}
  \vspace{-1em}
\end{figure}

In this section, we introduce a suite of training objectives (Fig.~\ref{Fig:training_pipeline}) to explicitly encourage \textbf{content preservation} and \textbf{instruction alignment}.
Our training pipeline integrates both supervised and unsupervised objectives, enabling editing models to benefit from paired data for precise editing control when available, while remaining 
applicable to large-scale, unpaired datasets.
In Section~\ref{sec:experiments}, we show that these enable state of the art instruction alignment and content preservation in driving scenes.

\vspace{-0.5em}

\subsubsection{Supervised Fine-Tuning for Instruction Alignment}
\label{sec:sft}

When paired source–target examples are available, we train image editing models with supervised fine-tuning (SFT).
Given paired training samples \((x_s, x_t, t_s, t_t, M_s, M_t)\), 
the generator model \(f_\theta\) produces an edited image, and we calculate the supervised fine-tuning loss in Eqn.~\ref{eqn:sft}.
  \vspace{-0.1cm}
\begin{equation}\label{eqn:sft}
\begin{aligned}
\hat{x}_t &= f_\theta(x_s, t_t, M_t), \\
\mathcal{L}_{\text{sft}} 
&= \lambda_{\text{sft}} \, \big\| x_t - \hat{x}_t \big\|_1
+ \lambda_{\text{sft-lpips}} \, \big\| \phi(x_t) - \phi(\hat{x}_t) \big\|_2,
\end{aligned}
\end{equation}
Each training instance includes a source scene ($x_s$), two natural language instructions describing the transformations from the source to the target ($t_t$), from target to the source ($t_s$), a LangMask for source to target transformations and vice versa ($M_t$, $M_s$ respectively), and the resulting edited target ($x_t$).
$\hat{x}_t = f_\theta(x_s, t_t, M_t)$ is the generator's output.
We use the Learned Perceptual Image Patch Similarity (LPIPS) loss \citep{zhang2018unreasonableeffectivenessdeepfeatures}, defined as the 
 distance between normalized deep features extracted from a pretrained VGG network $\phi(\cdot)$.

With this design, we can explicitly guide the model toward faithful instruction following. By conditioning the model on instruction-derived pixel masks, we constrain modifications to specified areas,
encouraging the model to localize edits.
This ensures unedited structures, such as road geometry and lane markings, are unchanged while surrounding vehicles and global features are edited.
Masked supervision penalizes deviations in non-edited regions, which also supports content preservation.

As a special case of $\mathcal{L}_\text{sft}$, when $x_s$ and $x_t$ are the same image (with a blank mask), we are essentially asking the editing model to preserve the content:

\begin{small}
  \vspace{-0.4cm}
\begin{align}
\mathcal{L}_{\text{id}} 
= {} & \lambda_{\text{id}} \, \big\| f_\theta(x_s, t_s, \emptyset) - x_s \big\|_1 \nonumber \\
& + \lambda_{\text{id-lpips}} \, \big\| \phi(f_\theta(x_s, t_s, \emptyset)) - \phi(x_s) \big\|_2
\vspace{-2cm}
\end{align}
\end{small}

This special case of $\mathcal{L}_\text{sft}$, i.e, an identity objective, enforces that when editing instructions correspond to no change (blank mask), the model reproduces the input.
This teaches the model to preserve dynamic scene instances that are not specified in the editing instruction.

\vspace{-0.5em}
\subsubsection{Language-Guided Cycle Consistency and Identity Preservation}
\label{sec:cycle_consistency}

While supervised fine-tuning provides precise control, acquiring paired data is costly, and the reliance on paired data limits its scalability to driving datasets where explicit ground truth edits are unavailable,
especially on unseen driving scenes in the wild where modern editing framework may fail.
Therefore, we choose to include complementary unsupervised constraints via cycle consistency and identity preservation, such that we can use them in out-of-distribution (OOD) unsupervised cases.

Cycle consistency extends this principle by encouraging reversibility. 
Without additional constraints, generative editing models may alter portions of the scene in regions unrelated to the instruction. 
By requiring the original image to be recoverable after a forward–backward editing cycle, the model is penalized for unnecessary deviations from the input.
This encourages content preservation by discouraging drift.
  \vspace{-0.2cm}
\begin{equation}
\begin{aligned}
\hat{x}_s &= f_\theta(f_\theta(x_s, t_t, M_t), t_s, M_s) \\
\mathcal{L}_{\text{cycle}} 
&= \lambda_{\text{cycle}} \, \big\| \hat{x}_s - x_s \big\|_1
+ \lambda_{\text{cycle-lpips}} \, \big\| \phi(\hat{x}_s) - \phi(x_s) \big\|_2.
\end{aligned}
\end{equation}

By combining pixel-level L1 and perceptual LPIPS losses, we enforce structural fidelity while allowing stylistic variation. 

The instructions for cycle consistency losses are sampled leveraging similar strategy as how we curate the pseudo editing pairs. More details can be found in Appendix~\ref{app:cycle-consistency-details}.

\vspace{-0.5em}
\subsubsection{Language-Guided CLIP Loss for Content Preservation}
\label{sec:clip_loss}

Reconstruction-based supervision is insufficient on its own: the model may collapse to an identity mapping, avoiding all edits to minimize loss. To overcome this degeneracy, we incorporate a complementary alignment signal based on language–image similarity, described in the this subsection.

\vspace{-0.4cm}
\begin{equation}
\label{eq:clip}
\begin{aligned}
\mathcal{L}_{\text{clip}} 
= {} & \lambda_{\text{clip}} \left( 1 - 
\operatorname{sim}_{\cos} \left( \text{CLIP}_I(\hat{x}_b), \, \text{CLIP}_T(t_b) \right) \right) \\
& + \lambda_{\text{clip}} \,
\operatorname{sim}_{\cos} \left( \text{CLIP}_I(\hat{x}_b), \, \text{CLIP}_T(t_a) \right),
\end{aligned}
  \vspace{-0.2cm}
\end{equation}
where $\text{CLIP}_I$ indicates CLIP's image feature, $\text{CLIP}_T$ is for CLIP's text feature, and $\operatorname{sim}_{\cos}$ for cosine similarity.
To counteract degraded generation, in Equation~\ref{eq:clip}, we first incorporate a language-guided CLIP similarity loss (the first term on right-hand side).
This loss measures the alignment between the generated output and the provided instruction using CLIP.
The aligned CLIP loss encourages outputs to move toward the intended semantic edit, thereby reinforcing instruction alignment. 
We also introduce a misalignment penalty term (the second term on right-hand side), which discourages similarity to input image description to prevent the model from reproducing the input.

Our final loss is: \begin{equation}\label{eq:final-loss}
\mathcal{L} 
= \mathcal{L}_{\text{sft}} 
+ \mathcal{L}_{\text{id}} 
+ \mathcal{L}_{\text{cycle}} 
+ \mathcal{L}_{\text{clip}} .
  \vspace{-0.2cm}
\end{equation}

\section{Experiments} \label{sec:experiments}

\subsection{Settings}

We evaluate our training strategies and pixel-level instructions on UltraEdit~\citep{zhao2024ultraeditinstructionbasedfinegrainedimage}. 
Following the evaluation protocol in \citep{zhang2024magicbrushmanuallyannotateddataset}, we assess editing performance using L1 distance, L2 distance, CLIP image similarity, and DINO similarity~\citep{caron2021emerging}. 
These metrics measure how well the edited image preserves the original content and reflects the required edit. \ourmethod trains on Boreas, and pseudo-pairs from Boreas \cite{burnett_ijrr23}, nuScenes \cite{caesar2020nuscenes}, and Argoverse2 \cite{wilson2023argoverse2}. We evaluate our models on Boreas to asses compound editing and nuPlan \cite{nuplan} to asses generalization over a distinct data distribution. Although nuPlan (test) and nuScenes (train) partially overlap in geographic coverage, we explicitly select nuPlan test images that are non-overlapping with any nuScenes training locations.

We structure our experiments to address our core questions in Section~\ref{core_questions}:
\textbf{Q1}: The role of \ourmethod in enabling intuitive, fine-grained editing at multiple levels of granularity (Sec.~\ref{sec:exp-general-performance}).
\textbf{Q2}: How fine-grained prompting promotes faithful instruction following while preserving scene content
(Sec.~\ref{sec:exp-ablation-guiding-driving-scene-editing}).
\textbf{Q3}: The importance of paired driving scene data and training objectives for generalizing across diverse unseen driving environments (Sec.~\ref{sec:exp-ablation-generalizable-editing-on-driving-scenes}).

\begin{figure*}[t!]
  \centering
  \vspace{-1em}
  \captionsetup{font=small}
  \includegraphics[clip, trim=0cm 0.1cm 0cm 0cm, width=0.8\textwidth]{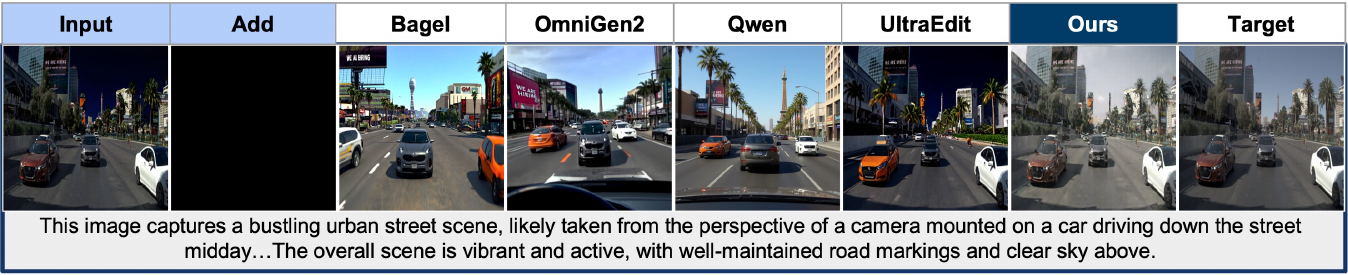}
  \includegraphics[clip, trim=0cm 0.1cm 0cm 0.8cm, width=0.8\textwidth]{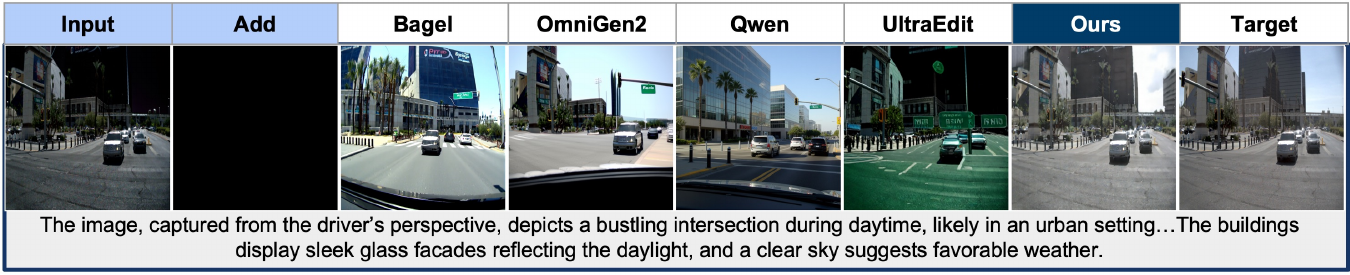}
  \includegraphics[clip, trim=0cm 0.1cm 0cm 0.1cm, width=0.8\textwidth]{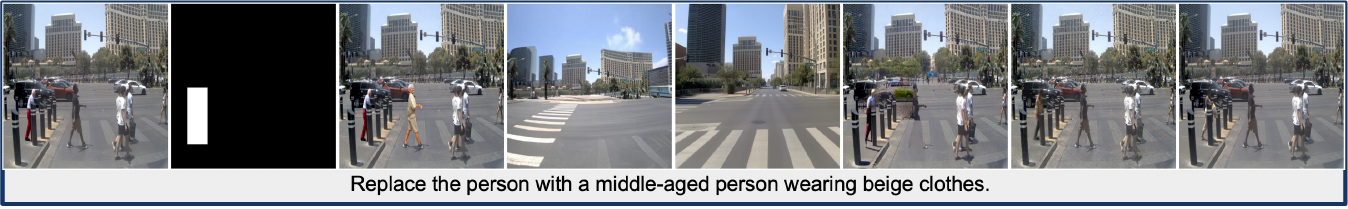}
  \includegraphics[clip, trim=0cm 0.1cm 0cm 0.2cm, width=0.8\textwidth]{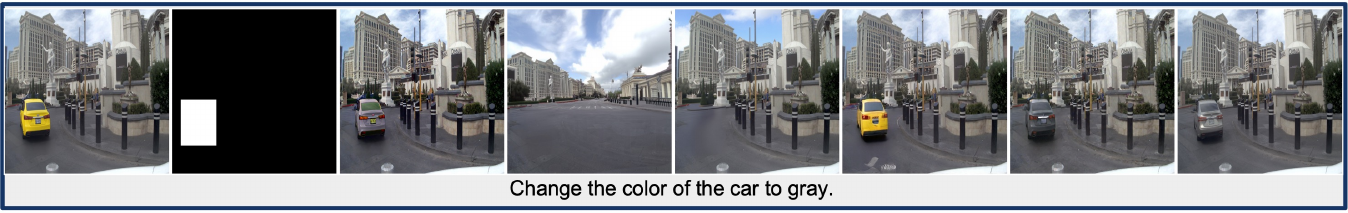}
  \includegraphics[clip, trim=0cm 0.2cm 0cm 0.8cm, width=0.8\textwidth]{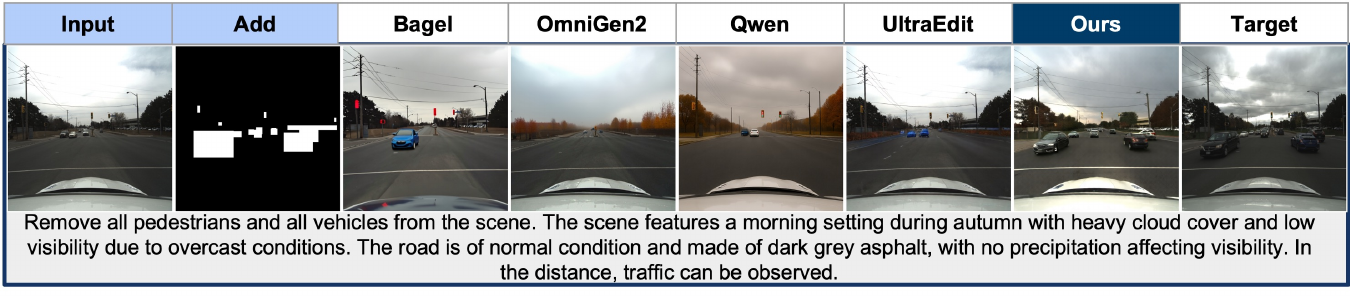}
  \includegraphics[clip, trim=0cm 0cm 0cm 0.8cm, width=0.8\textwidth]{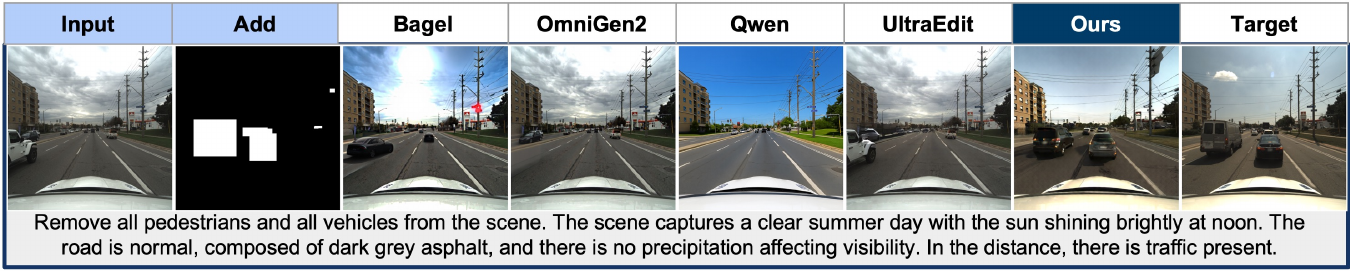}
  \captionsetup{font=small}
  \vspace{-0.7em}
  \caption{\textbf{\ourmethod Editing.} Rows 1,2 Local edits: the masks (projected as binary images and stated in text for reference) enable modifications to traffic. Rows 3, 4 Global edits: the text prompt informs the appearance of the scene. For brevity, only the portions relevant to the shown edits are displayed. Rows 5, 6 Compound edits: The masks (projected as binary images) enable modifications to traffic while the text prompt informs the desired global appearance. We compare to Qwen~\cite{wu2025qwenimagetechnicalreport}, OmniGen2~\cite{wu2025omnigen2}, UltraEdit~\cite{zhao2024ultraeditinstructionbasedfinegrainedimage}, and BAGEL~\cite{deng2025emergingpropertiesunifiedmultimodal}.}\label{exp:fig-all-editing}
  \vspace{-1em}
\end{figure*}

\begin{table*}[thb]
\caption{\textbf{Fine-grained, Global, and OOD Editing Results.} Each model is evaluated across fine-grained (full image), fine-grained (crop), and global editing tasks on nuPlan which is outside of the training dataset distribution. We report local-editing performance both over the \textit{full image} and an edited-region \textit{crop} to evaluate both content preservation and multi-scale instruction alignment.}
\vspace{-0.5em}
\centering
\small
\setlength{\tabcolsep}{3pt}
\renewcommand{\arraystretch}{0.85}
\begin{tabular}{lcccccccccccc}
\toprule
& \multicolumn{4}{c}{\textbf{Fine-Grained (Full Image)}} 
& \multicolumn{4}{c}{\textbf{Fine-Grained (Crop)}} 
& \multicolumn{4}{c}{\textbf{Global}} \\
\cmidrule(lr){2-5} \cmidrule(lr){6-9} \cmidrule(lr){10-13}
\textbf{Model} 
& \textbf{L1} ↓ & \textbf{L2} ↓ & \textbf{CLIP} ↑ & \textbf{DINO} ↑ 
& \textbf{L1} ↓ & \textbf{L2} ↓ & \textbf{CLIP} ↑ & \textbf{DINO} ↑
& \textbf{L1} ↓ & \textbf{L2} ↓ & \textbf{CLIP} ↑ & \textbf{DINO} ↑ \\
\midrule
Bagel    
& 0.1548 & 0.0600 & 0.8381 & 0.7200
& 0.1645 & 0.0569 & 0.7498 & 0.5075
& 0.2925 & 0.1370 & 0.7952 & 0.6361 \\
OmniGen2 
& 0.2406 & 0.1109 & 0.7689 & 0.5539
& 0.2039 & 0.0735 & 0.7043 & 0.3383
& 0.2797 & 0.1250 & 0.7908 & 0.6365 \\
Qwen     
& 0.1577 & 0.0581 & 0.8387 & 0.7302
& 0.1505 & 0.0463 & 0.7688 & 0.5289
& 0.2618 & 0.1136 & 0.8196 & 0.7073 \\
UltraEdit
& 0.0341 & 0.0047 & 0.9278 & 0.8993
& 0.0964 & 0.0239 & 0.7451 & 0.4791
& 0.1881 & 0.0635 & 0.8086 & 0.7374 \\
\textbf{Ours} 
& \textbf{0.0301} & \textbf{0.0033} & \textbf{0.9473} & \textbf{0.9298}
& \textbf{0.0727} & \textbf{0.0164} & \textbf{0.8113} & \textbf{0.6563}
& \textbf{0.1025} & \textbf{0.0197} & \textbf{0.9189} & \textbf{0.9110} \\
\bottomrule
\end{tabular}
\label{tab:combined_OOD_editing_performance}
\vspace{-1em}
\end{table*}

\subsection{\ourmethod: Fine-Grained Scene Editing}\label{sec:exp-general-performance}

The careful design of \ourmethod across data (Section~\ref{sec:automatic-paired-dataset-generation}), model (Section~\ref{sec:LangMasks for fine-grained editing}), and training (Section~\ref{sec:methods-training-objectives}) jointly leads to state-of-the-art instruction alignment and content preservation. \textbf{Targeting Q1, our evaluation includes three levels of granularity: local editing, global editing, and compound editing.} Compound editing is evaluated with real-world image pairs from Boreas, while isolated local and global edits are evaluated over unseen pseudo-pairs from nuPlan. The latter additionally allows assessing generalization robustness (Q3) in more detail in Sec.~\ref{sec:exp-ablation-generalizable-editing-on-driving-scenes}.
Tab.~\ref{tab:combined_OOD_editing_performance} shows that our model achieves the \textbf{lowest L1 and L2 errors} across both local and global edits, demonstrating strong content preservation in unedited areas and high instruction fidelity in edited areas. 
Tab.~\ref{tab:compound_editing_performance} shows that \ourmethod remains competitive with Qwen, despite using a 7B UltraEdit backbone vs. Qwen’s 20B, and achieves the \textbf{highest CLIP and DINO similarity} across all granularities reflecting superior instruction alignment while preserving scene content. 
We report local-editing performance both over the full image (content preservation + alignment) and an edited-region crop (fine-grained alignment on small objects).

The qualitative results in Fig.~\ref{exp:fig-all-editing} shows our model can simultaneously perform precise editing at local (rows 3-4) and global scales (rows 1,2), and compound editing (rows 5–6). In contrast, baseline models may fail to preserve critical scene elements such as buildings or fail to follow the instruction. For fairness, all baseline models have the LangMask instructions appended to their global text prompts. UltraEdit, supporting binary masks, additionally receives the projected binary LangMask.

\begin{table}[h]
\caption{\textbf{Compound Editing.} We evaluate across real-world local and global edits simultaneously.}
\vspace{-0.8em}
\centering
\small
\setlength{\tabcolsep}{4pt}
\renewcommand{\arraystretch}{0.85}
\begin{tabular}{lcccc}
\toprule
\textbf{Model} & \textbf{L1} (↓) & \textbf{L2} (↓) & \textbf{CLIP} (↑) & \textbf{DINO} (↑) \\
\midrule
Bagel                        & 0.2258 & 0.0899 & 0.8224 & 0.6918 \\
OmniGen2                     & 0.2313 & 0.0948 & 0.8385 & 0.7212 \\
Qwen                         & \textbf{0.1868} & \textbf{0.0603} & 0.8640 & 0.7896 \\
UltraEdit                    & 0.2208 & 0.0848 & 0.8667 & 0.7838 \\
\textbf{Ours}    & 0.1955 & 0.0708 & \textbf{0.8740} & \textbf{0.8232} \\
\bottomrule
\end{tabular}
\label{tab:compound_editing_performance}
\vspace{-2em}
\end{table}

\subsection{Promoting Instruction Alignment and Content Preservation}\label{sec:exp-ablation-guiding-driving-scene-editing}
To understand how our method promotes instruction alignment and content preservation, we evaluate it using both ablation studies and human assessments.
In Tab.~\ref{tab:ablation-lang-mask}, we carefully ablate our method over our real-world compound editing dataset. We train our base model, UltraEdit, to employ text prompts (for both local and global changes) with binary masks for localized edits (``UltraEdit-Text''); then a model that employs LangMasks with global prompts (``UltraEdit-Mask''). When we add our unsupervised training methods (``UltraEdit-Mask-UFT'') we see a significant improvement in all scores suggestion higher instruction alignment in edited regions and better content preservation in unedited regions.

\begin{table}[H]
\vspace{-1em}
\caption{\textbf{LangMasks and Training Objectives.} Masks describing pixel level changes and unsupervised training objectives promote instruction alignment and content preservation on compound edits.}
\vspace{-0.3cm}
\centering
\small
\setlength{\tabcolsep}{4.5pt}
\renewcommand{\arraystretch}{0.85}
\resizebox{0.45\textwidth}{!}{
\begin{tabular}{lcccc}
\toprule
\textbf{Model} & \textbf{L1} (↓) & \textbf{L2} (↓) & \textbf{CLIP} (↑) & \textbf{DINO} (↑) \\
\midrule
UltraEdit                    & 0.2208 & 0.0848 & 0.8667 & 0.7838 \\
UltraEdit-Text           & 0.2324 & 0.0996 & 0.8565 & 0.7323 \\
UltraEdit-Mask           & 0.2278 & 0.0880 & 0.8726 & 0.7849 \\
UltraEdit-Mask-UFT & \textbf{0.1733} & \textbf{0.0540} & \textbf{0.8758} & \textbf{0.8003} \\
\bottomrule
\end{tabular}
}
\label{tab:ablation-lang-mask}
\vspace{-0.2cm}
\end{table}

\subsubsection{User Study: Evaluating Instruction Alignment and Content Preservation}

To assess the instruction fidelity and content preservation (\textbf{Q2}) of \ourmethod, we perform a user study with 26 participants (409 responses). Each participant is shown multiple randomly selected example from our local, global, and compound editing evaluation sets. For each example, participants view one output from each model alongside the global prompt, the instance level prompt, and the projected binary LangMask and are asked to select the image that best \textbf{follows the editing instruction while preserving unedited regions and maintaining overall image quality and realism}. As shown in Tab.~\ref{table:user-study}, \ourmethod is \textbf{overwhelmingly preferred achieving a 61.1\% preference rate} and \textbf{61\% win rate}, far exceeding all baselines. Users consistently favored our results for their balance of faithful edits and high scene realism.

\begin{table}[H]
\vspace{-0.3cm}
\centering
\footnotesize
\caption{\textbf{User study results}: overall preference distribution (``Pref.'' in \%) and win rates including local, global, and compound edits.}
\vspace{-0.3cm}
\label{table:user-study}
\begin{tabular}{lccccc}
\toprule
Model & Bagel & Omnigen2 & Qwen & UltraEdit & \textbf{Ours} \\
\midrule
Pref. (\%) & 9.0 & 2.0 & 14.7 & 13.2 & \textbf{61.1} \\
Win        & 9.0 & 2.0 & 14.7 & 13.2 & \textbf{61.3} \\
\bottomrule
\end{tabular}
\vspace{-0.5em}
\end{table}

\subsection{Generalizable Editing on Driving Scenes}\label{sec:exp-ablation-generalizable-editing-on-driving-scenes}

To evaluate \textbf{Q3}, we test \ourmethod on OOD datasets. Although nuPlan (test) and nuScenes (train) partially overlap in geographic coverage, we explicitly select nuPlan test images that are non-overlapping with any nuScenes training locations. In addition to the state-of-the-art editing performance on the unseen nuPlan dataset (Tab.~\ref{tab:combined_OOD_editing_performance}), Tab.~\ref{tab:ablation-pseudo-image-pairs} shows that adding pseudo-image pairs meaningfully improves generalization.
\ourmethod shows the best content preservation and instruction alignment compared to a model trained with only our real-world image pairs using LangMasks and our training objectives when editing unseen driving scenes.

\begin{table}[H]
\vspace{-1em}
\caption{\textbf{Diverse Environment Editing.} Each model is evaluated across fine-grained (full image), fine-grained (crop), and global editing tasks on nuPlan.}
\vspace{-0.5em}
\centering
\scriptsize
\setlength{\tabcolsep}{4pt}
\renewcommand{\arraystretch}{0.8}

\begin{tabular}{llcccc}
\toprule
\textbf{Setting} & \textbf{Model} & \textbf{L1} (↓) & \textbf{L2} (↓) & \textbf{CLIP} (↑) & \textbf{DINO} (↑) \\
\midrule

\multirow{2}{*}{Full Image}
& UltraEdit-Mask-UFT 
& 0.1851 & 0.0585 & 0.7464 & 0.5521 \\
& \textbf{Ours} 
& \textbf{0.0301} & \textbf{0.0033} & \textbf{0.9473} & \textbf{0.9298} \\

\midrule

\multirow{2}{*}{Crop}
& UltraEdit-Mask-UFT 
& 0.1858 & 0.0616 & 0.7001 & 0.3071 \\
& \textbf{Ours} 
& \textbf{0.0727} & \textbf{0.0164} & \textbf{0.8113} & \textbf{0.6563} \\

\midrule

\multirow{2}{*}{Global}
& UltraEdit-Mask-UFT 
& 0.2534 & 0.1067 & 0.7278 & 0.5560 \\
& \textbf{Ours} 
& \textbf{0.1025} & \textbf{0.0197} & \textbf{0.9189} & \textbf{0.9110} \\

\bottomrule
\end{tabular}

\label{tab:ablation-pseudo-image-pairs}
\vspace{-1em}
\end{table}

\subsubsection{Safety-Critical Downstream Applications}

Generalization across domains is essential for safety-critical autonomy pipelines.
Using our model, we synthesize images from realistic images taken from one random quarter of the nuScenes dataset under various weather conditions. Then we augment the original quarter dataset with these images and train two BEV map segmentation models with and without the synthesized images. Following~\citep{liu2022bevfusion}, we report the best per-class IoU in Tab. \ref{table:EXP-downstream-task-example}. This augmentation improves mean IoU by 33\% in real-world scenes, showing that \ourmethod provides scene variation that strengthens downstream perception.

\begin{table}[H]
\vspace{-0.3cm}
\centering
\footnotesize
\caption{\textbf{BEV Map Segmentation}. Intersection-over-Union (IoU) across 6 classes and the class-averaged IoU on nuScenes.}
\vspace{-0.3cm}
\label{table:EXP-downstream-task-example}
\renewcommand{\arraystretch}{0.85}
\resizebox{0.47\textwidth}{!}{
\begin{tabular}{lccccccc}
\toprule
{} & Drivable & Ped.~Cross. & Walkway & Stop~Line & Carpark & Divider & \textbf{Mean} \\
\midrule
Original   & 0.5834 & 0.0533 & 0.1626 & 0.0609 & 0.0981 & 0.1569 & \textbf{0.1859} \\
Augmented  & 0.6448 & 0.1147 & 0.2251 & 0.1059 & 0.1776 & 0.2180 & \textbf{0.2477} \\
\bottomrule
\end{tabular}
}
\vspace{-0.2cm}
\end{table}

\section{Conclusion}
In this work, we present \ourmethod, a scalable framework for instruction-guided editing of complex
driving scenes that unifies advances in data, model, and training. It tackles three key challenges—multi-level granularity, rich semantics, and robustness to domain shifts—while achieving strong instruction alignment, content preservation, and generalization.
Limitations: Although \ourmethod relies on LLM-generated instructions and VLM-based filtering, which may introduce hallucinations and noisy pseudo-pairs, we mitigate these with algorithmic/VLM filters and point to verification and human-in-the-loop strategies as future work.

{
    \small
    \bibliographystyle{ieeenat_fullname}
    \bibliography{main}
}

\clearpage
\setcounter{page}{1}
\maketitlesupplementary

\section{Dataset Collection Details}\label{app_sec:dataset-collection-details}
\subsection{Real-World Data Pairing Details}\label{app_sec:real-world-data-pairing-details}
Given a multi-season driving dataset with repeated routes and calibrated camera poses, we convert unpaired recordings into pose-aligned image pairs using a simple geometric matching rule.

Let $I_{\text{source}}$ be a frame with camera pose $(\vec{x}_s, \phi_s, \theta_s, \psi_s)$, where $\vec{x}_s \in \mathbb{R}^3$ is the camera position and $(\phi_s, \theta_s, \psi_s)$ denote roll, pitch, and yaw. We first define a local temporal neighborhood $\mathcal{I}$ of candidate frames from other traversals of the same route. The paired target frame $I_{\text{target}}$ is selected by minimizing a pose-disparity distance:
\begin{equation}\label{eqn:min-camera-pose}
\begin{aligned}
I_{\text{target}} &= \arg\min_{I \in \mathcal{I}} \; \text{dist}(I, I_{\text{source}}), \\
\text{dist}(I_a, I_b) &= 
\bigl\lVert \vec{x}_a - \vec{x}_b \bigr\rVert_2
+ \bigl| \phi_a - \phi_b \bigr|
+ \bigl| \theta_a - \theta_b \bigr| \\
&~\quad+ \bigl| \psi_a - \psi_b \bigr|,
\end{aligned}
\end{equation}
where $(\vec{x}_a, \phi_a, \theta_a, \psi_a)$ and $(\vec{x}_b, \phi_b, \theta_b, \psi_b)$ are the poses of frames $I_a$ and $I_b$, respectively. In practice, we accept matches only if $\text{dist}(I_{\text{target}}, I_{\text{source}})$ falls below a fixed threshold, ensuring that the resulting pairs share nearly identical viewpoints. This pose-based pairing procedure is dataset-agnostic and can be applied to any driving dataset with sufficiently accurate pose estimates, including but not limited to Boreas.

\subsection{Image Descriptor} \label{app_sec:image-descriptor}
Our ``Image Descriptor'', an image-based analogue of iFinder~\cite{yao2025ifinder}, is a \textbf{comprehensive annotation system}: it integrates \ul{vision-language and depth estimation models} that can generate semantic-aligned scene descriptions at two levels:

\mypara{Multimodal Environments Descriptions}
We first extract global information about the scene:
\begin{enumerate}[leftmargin=*]
    \item We use an image‑based vision‑language model (VLM)~\citep{chen2024internvl} for a global interpretation of extremely fine-grained attributes. We show the VLM prompt in Appendix \ref{app:sec-example-annotation}.
    \item To estimate object distances, we apply a metric depth estimation model, Metric3d~\citep{hu2024metric3dv2}, to the full image, producing a depth map whose values correspond to real‑world distances.
\end{enumerate}

\mypara{Instance-Level Semantic Decomposition}
After preparing the global description, we then record objects present in the scene:
\begin{enumerate}[leftmargin=*]
    \item We run a 2D object detector (Owlv2~\citep{minderer2024scalingopenvocabularyobjectdetection}) that returns, for each detected object, a bounding box, a class label (from the set `ambulance', `bicycle', `traffic light', `traffic cone', `person', `car', `motorcycle', `bus', `building', `fire truck'), and a unique object ID.
    \item For each object, we crop the global depth map (from the 2nd step above) to its bounding box and then refine that region with a binary mask from the Segment Anything Model (SAM~\citep{kirillov2023segment-anything}), ensuring we exclude background pixels. The object’s distance is taken as the mean depth over this masked area.
    \item We invoke the VLM~\citep{chen2024internvl} on each object’s bounding box to extract additional attributes, such as vehicle color or traffic‑light state.
\end{enumerate}
We show an example annotation in Sec \ref{app:sec-example-annotation}.

\subsection{Global Editing Details}\label{app_sec:global-editing-details}

We define three categories of global scene edits that capture the dominant real\mbox{-}world variations in driving environments:
\begin{itemize}
    \item \textbf{Weather:} \emph{Sunny, Cloudy, Foggy, Rainy, Snowy}
    \item \textbf{Time of day:} \emph{Dawn, Day, Dusk, Night}
    \item \textbf{Season:} \emph{Spring, Summer, Autumn, Winter}
\end{itemize}
Together, these attributes span a broad range of environmental conditions and naturally occurring appearance changes. Applying such global edits to driving scenes produces diverse variations of the same location, which is crucial for evaluating the robustness and generalization capability of autonomous driving systems.

\subsection{Quality-control pipeline for local pseudo-edited samples.}\label{app_sec:quality-control-local-edit}
We perform automated quality control using a combination of a Vision-Language Model (VLM) and image-based similarity metrics to remove local low-quality pseudo-edits.

\mypara{Global sanity check}
Given an edited image, we first prompt the VLM to flag samples that appear clearly synthetic or contain obvious artifacts (e.g., distorted geometry, inconsistent lighting). Samples flagged as unrealistic are discarded.

\mypara{Pedestrian edits}
For edits involving pedestrians, we crop and enlarge each modified person and query the VLM with a category-specific prompt asking whether the subject looks realistic after modification or replacement. Crops judged unrealistic are removed from the dataset.

\mypara{Vehicle edits}
For vehicles, we similarly crop each edited instance and prompt the VLM to predict the vehicle’s orientation (e.g., facing forward, turning left). We then check whether this predicted orientation is consistent with the intended edit and the surrounding context. Samples with mismatched or ambiguous orientation are discarded.

\mypara{Object removals}
For removal edits, we extract enlarged crops from both the original and edited images corresponding to the target region. The VLM is asked whether an instance of the removed class is still visible in the edited crop. If the VLM indicates successful removal, we additionally compute the Structural Similarity Index (SSIM) between the Canny edge maps of the original and edited crops. Only removals that pass both the semantic (VLM) and structural (edge-based SSIM) checks are retained, which is particularly important for densely packed objects such as vehicles.

\mypara{Traffic light edits}
For traffic lights, we restrict edits to color changes. A valid edit must satisfy three conditions:
(i) the VLM confirms the presence of a plausible traffic light in the edited crop;
(ii) the predicted signal color matches the target color specified by the edit;
(iii) the SSIM between the edge maps of the original and edited crops remains above a fixed threshold, indicating preserved structure of the traffic light and its surroundings.

This quality-control procedure is dataset-agnostic and can be applied to any pseudo-edited driving dataset to systematically filter out unrealistic or structurally inconsistent edits.

\section{Dataset Statistics}

\subsection{Compound Editing}

In Tab.~\ref{table:real-dataset-statistics} we show the editing types in the compound editing dataset derived from the real-world dataset Boreas~\cite{burnett_ijrr23}. 
In Tab.~\ref{table:dataset_compound_subedit_counts}, we detail how many edit types each example contains.

\begin{table}[H]
\centering
\captionsetup{font=small}
\caption{\textbf{Compound Dataset Statistics (Edit Types).}
Distribution of edit types across the compound editing dataset.}
\label{table:real-dataset-statistics}
\begin{tabular}{lrr}
\hline
\textbf{Edit Type} & \textbf{Count} & \textbf{Percentage} \\
\hline
Road Conditions & 146{,}159 & 15\% \\
Time of Day     & 235{,}710 & 24\% \\
Traffic         & 263{,}064 & 27\% \\
Traffic Light   & 55{,}998  & 6\% \\
Weather         & 269{,}715 & 28\% \\
\hline
\end{tabular}
\vspace{-0.5em}
\end{table}

\begin{table}[H]
\centering
\captionsetup{font=small}
\caption{\textbf{Compound Dataset Statistics (Subedit Counts).}
Distribution of the number of edit types per example of the compound editing dataset.}
\label{table:dataset_compound_subedit_counts}
\begin{tabular}{lrr}
\hline
\multicolumn{3}{l}{\textbf{Partition by Number of Subedits}} \\
\hline
1 Edit Type  & 16{,}623  & 4.86\% \\
2 Edit Types & 95{,}295  & 27.88\% \\
3 Edit Types & 160{,}413 & 46.93\% \\
4 Edit Types & 64{,}385  & 18.84\% \\
5 Edit Types & 4{,}922   & 1.44\% \\
\hline
\textbf{Total Examples} & 341{,}796 & -- \\
\hline
\end{tabular}
\vspace{-0.5em}
\end{table}

\subsection{Local Editing}

We breakdown the local editing pseudo-pair datasets by generated editing type and targeted subject class for nuScenes (Tab.~\ref{app:table-nuscenes-local-stats}), Boreas (Tab.~\ref{app:table-boreas-local-stats}), Argoverse2 (Tab.~\ref{app:table-argoverse-local-stats}), and nuPlan (Tab.~\ref{app:table-nuplan-local-stats}).

\begin{table}[h]
\centering
\captionsetup{font=small}
\caption{\textbf{nuScenes — Editing Statistics.} Editing types and editing actions for the nuScenes dataset.}
\label{app:table-nuscenes-local-stats}
\begin{tabular}{l r c l r}
\hline
\textbf{Editing Type} & \textbf{Count} && \textbf{Action} & \textbf{Count} \\
\hline
vehicle      & 3{,}584 && modify        & 2{,}035 \\
pedestrian   & 1{,}879 && replace       & 1{,}559 \\
trafficlight & 41      && delete/insert & 1{,}910 \\
\hline
\textbf{Total} & 5{,}504 \\
\hline
\end{tabular}
\vspace{-0.5em}
\end{table}

\begin{table}[h]
\centering
\captionsetup{font=small}
\caption{\textbf{Boreas — Editing Statistics.} Editing types and editing actions for the Boreas dataset.}
\label{app:table-boreas-local-stats}
\begin{tabular}{l r c l r}
\hline
\textbf{Editing Type} & \textbf{Count} && \textbf{Action} & \textbf{Count} \\
\hline
vehicle      & 5{,}729 && modify        & 2{,}070 \\
pedestrian   & 96      && delete/insert & 2{,}957 \\
trafficlight & 144     && replace       & 942 \\
\hline
\textbf{Total} & 5{,}969  \\
\hline
\end{tabular}
\vspace{-0.5em}
\end{table}

\begin{table}[h]
\centering
\captionsetup{font=small}
\caption{\textbf{Argoverse — Editing Statistics.} Editing types and editing actions for the Argoverse dataset.}
\label{app:table-argoverse-local-stats}
\begin{tabular}{l r c l r}
\hline
\textbf{Editing Type} & \textbf{Count} && \textbf{Action} & \textbf{Count} \\
\hline
vehicle      & 2{,}558 && modify        & 1{,}815 \\
trafficlight & 229    && replace       & 1{,}503 \\
pedestrian   & 2{,}031 && delete/insert & 1{,}500 \\
\hline
\textbf{Total} & 4{,}818  \\
\hline
\end{tabular}
\vspace{-0.5em}
\end{table}

\begin{table}[h]
\centering
\captionsetup{font=small}
\caption{\textbf{nuPlan — Editing Statistics.} Editing types and editing actions for the nuPlan dataset.}
\label{app:table-nuplan-local-stats}
\begin{tabular}{l r c l r}
\hline
\textbf{Editing Type} & \textbf{Count} && \textbf{Action} & \textbf{Count} \\
\hline
vehicle      & 50 && delete/insert & 33 \\
trafficlight & 23 && replace       & 31 \\
pedestrian   & 48 && modify        & 57 \\
\hline
\textbf{Total} & 121 \\
\hline
\end{tabular}
\vspace{-0.5em}
\end{table}

\subsection{Global Editing}

We breakdown the global editing pseudo-pair datasets for nuScenes (Tab.~\ref{app:table:nuscenes-global-edit-types}), Argoverse2 (Tab.~\ref{app:table:argoverse-global-edit-types}), and nuPlan (Tab.~\ref{app:table:nuplan-global-edit-types}).

\begin{table}[H]
\centering
\captionsetup{font=small}
\caption{\textbf{nuScenes Global Editing Type.}
Distribution of global editing types in the nuScenes dataset.}
\label{app:table:nuscenes-global-edit-types}
\begin{tabular}{lrr}
\hline
\textbf{Edit Type} & \textbf{Count} & \textbf{Percentage} \\
\hline
Season        & 3{,}033 & 33\% \\
Weather       & 3{,}116 & 33\% \\
Time of Day   & 3{,}162 & 34\% \\
\hline
\textbf{Total} & 9{,}311 & -- \\
\hline
\end{tabular}
\vspace{-0.5em}
\end{table}

\begin{table}[H]
\centering
\captionsetup{font=small}
\caption{\textbf{Argoverse2 Global Editing Types.}
Distribution of global editing types in the Argoverse2 dataset.}
\label{app:table:argoverse-global-edit-types}
\begin{tabular}{lrr}
\hline
\textbf{Edit Type} & \textbf{Count} & \textbf{Percentage} \\
\hline
Season           & 3{,}267 & 33\% \\
Weather          & 3{,}288 & 33\% \\
Time of Day      & 3{,}409 & 34\% \\
\hline
\textbf{Total}   & 9{,}964 & -- \\
\hline
\end{tabular}
\vspace{-0.5em}
\end{table}

\begin{table}[H]
\centering
\captionsetup{font=small}
\caption{\textbf{nuPlan Global Editing Types.}
Distribution of global editing types in the nuPlan dataset.}
\label{app:table:nuplan-global-edit-types}
\begin{tabular}{lrr}
\hline
\textbf{Edit Type} & \textbf{Count} & \textbf{Percentage} \\
\hline
Season        & 112 & 35\% \\
Weather       & 100 & 31\% \\
Time of Day   & 108 & 34\% \\
\hline
\textbf{Total} & 320 & -- \\
\hline
\end{tabular}
\vspace{-0.5em}
\end{table}

\section{Annotation Pipeline}\label{App:sec-annotation-pipeline}

\begin{figure*}[h]
\centering
\begin{minipage}{0.90\textwidth}
\lstset{
    basicstyle=\footnotesize\ttfamily,
    breaklines=true,
    breakatwhitespace=true,
    postbreak=\mbox{\textcolor{gray}{$\hookrightarrow$}\space},
}
\begin{lstlisting}
You are an expert in autonomous driving, specializing in analyzing traffic scenes. You receive a series of traffic images from the perspective of the ego car. Your task is to describe the driving environment, focusing on weather, lighting, road layout, and environment.

It is essential that you strictly follow the rules and instructions below. Any deviation from the specified structure or format will result in an invalid output.

STRICTLY follow Rules:
 - You must strictly follow the dictionary structure provided above.
 - Only use the specified terms for weather, light, road layout, and environment. Do not create your own terms.
 - No additional information or categories should be added.
 - You should strictly follow these instructions. If an object or element is not visible or does not exist in the scene, set the value to 'None'. Ensure every field is filled with the appropriate value or 'None'.
 - STRICTLY ignore any text written on the image. 


Output the result in the following dictionary format:

{
  "surrounding_info": {
    "weather": "[e.g., 'cloudy', 'sunny', 'rainy', 'fog', 'snowy']",
    "road_layout": "[Choose from: 'straight road', 'curved road', 'intersection', 'T-junction', 'ramp']",
    "environment": "[Choose from: 'city street', 'country road', 'highway', 'residential area']",
    "sun_visibility_conditions": "[Choose from: 'clear', 'foggy', 'low visibility', 'hazy']",
    "road_condition": "[Choose from: 'wet', 'icy', 'normal', 'debris', 'potholes']",
    "surface_type": "[Choose from: 'asphalt', 'gravel', 'dirt', 'concrete']",
    "surface_color": "[Choose from: 'light grey', 'dark grey', 'black', 'brown']",
    "time_of_the_day": "[Choose from: 'morning', 'midday', 'afternoon', 'night', 'dawn', 'dusk'.]",
    "precipitation_intensity": "[Choose from: 'none', 'light', 'moderate', 'heavy', 'torrential'.]",
    "precipitation_visibility_impact": "[Choose from: 'none', 'low', 'moderate', 'high']",
    "cloud_cover": "[Choose from: 'clear', 'light', 'moderate', 'heavy'.]
    }
}
\end{lstlisting}
\end{minipage}
\caption{Annotation pipeline prompt used for the VLM.}\label{App:annotation-pipeline-prompt}
\end{figure*}

\subsection{Example Annotation}\label{app:sec-example-annotation}

\begin{figure*}[h]
\vspace{-2em}
\centering
\includegraphics[width=0.75\textwidth]{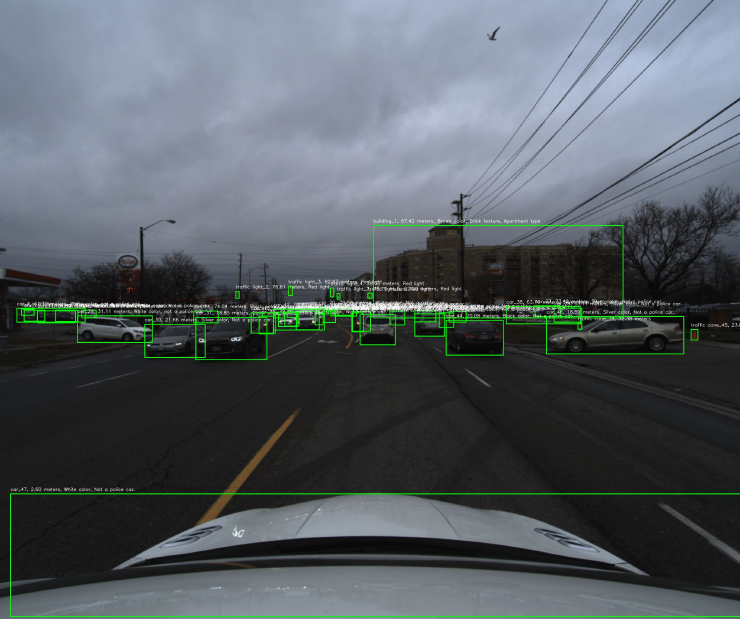}
\caption{Example annotation produced by the annotation pipeline.}
\vspace{-2em}
\label{app:fig-example-annotated-image}
\end{figure*}

The VLM is prompted with the instruction shown in Fig. \ref{App:annotation-pipeline-prompt}.
We show an annotated image (Fig.~\ref{app:fig-example-annotated-image}) and a truncated output caption from the annotation pipeline (Fig.~\ref{app:fig-example-annotation-truncated}).

\begin{figure*}[h]
\centering
\begin{minipage}{0.90\textwidth}

\lstset{
    basicstyle=\footnotesize\ttfamily,
    breaklines=true,
    breakatwhitespace=true,
    postbreak=\mbox{\textcolor{gray}{$\hookrightarrow$}\space},
}
\begin{lstlisting}
{
    "surrounding_info": {
        "weather": "cloudy",
        "road_layout": "intersection",
        "environment": "city street",
        "sun_visibility_conditions": "low visibility",
        "road_condition": "normal",
        "surface_type": "asphalt",
        "surface_color": "dark grey",
        "time_of_the_day": "morning",
        "precipitation_intensity": "none",
        "precipitation_visibility_impact": "none",
        "cloud_cover": "heavy"
    },
    "object_info": [
        {
            "class": "building",
            "bbox": [
                1234,
                745,
                2060,
                1051
            ],
            "object_id": 1,
            "distance_from_ego_vehicle": "67.42 meters",
            "attributes": "Brown color, Brick texture, Apartment type"
        },
        {
            "class": "traffic light",
            "bbox": [
                779,
                963,
                790,
                986
            ],
            "object_id": 2,
            "distance_from_ego_vehicle": "78.61 meters",
            "attributes": "Red light"
        },
        {
            "class": "car",
            "bbox": [
                79,
                1025,
                147,
                1062
            ],
            "object_id": 10,
            "distance_from_ego_vehicle": "69.81 meters",
            "attributes": "White color, Not a police car."
        },
        ...
    ]
}
\end{lstlisting}
\end{minipage}
\caption{Truncated annotation output corresponding to Fig.~\ref{app:fig-example-annotated-image}.\label{app:fig-example-annotation-truncated}}
\end{figure*}

\subsection{LLM Generated Editing Instructions}\label{App:real-gpt-prompt}

We prompt chatGPT-4o mini with the following instruction (Fig.~\ref{App:real-gpt-prompt-listing}) to produce editing instructions for real-world images based on the captions of the target image.

\begin{figure*}[h]
\centering
\begin{minipage}{0.90\textwidth}

\lstset{
    basicstyle=\footnotesize\ttfamily,
    breaklines=true,
    breakatwhitespace=true,
    postbreak=\mbox{\textcolor{gray}{$\hookrightarrow$}\space},
}
\begin{lstlisting}
You are an expert in autonomous driving, specializing in analyzing traffic scenes. You receive a text description of a traffic image from the perspective of an autonomous vehicle's camera.

Your task is to produce FOUR VERSIONS of the SAME PROMPT, each with DIFFERENT WORDING BUT IDENTICAL CONTENT, that describes the driving scene depicted in the image.

IMPORTANT: 
- Each version should contain the EXACT SAME DESCRIPTION, just phrased differently.
- ALL prompts should describe EXACTLY THE SAME SCENE with no variation in what is being described.
- Only use adjectives and descriptors that are explicitly provided in the caption. Do NOT add your own subjective descriptors like "moody," "tranquil," "charming," etc. Stick strictly to the attributes and descriptors that appear in the input caption.

At the end of each prompt version, append the following line:
    "There may be minor additional changes in time or weather (such as lighting, clouds, or rain) between the images that are not fully captured by the descriptions, but these are expected to be subtle."

The prompt should be in the format below where each version describes the same contents but with different wording. 

### Scene Description:

version_1: {{description_1}}
version_2: {{description_2}}
version_3: {{description_3}}
version_4: {{description_4}}

Image Caption: {caption_0}
\end{lstlisting}
\end{minipage}
\caption{Prompt used to generate real-world image editing instructions.}
\label{App:real-gpt-prompt-listing}
\end{figure*}

\section{Efficient LangMask Construction from User Instructions}\label{app_sec:langMasks}

To enable precise, instruction-aligned scene editing, we provide two efficient mechanisms for constructing LangMasks from user-specified edits.

\mypara{Dataset-Driven LangMask Generation} First, following \ourmethod's dataset construction, a driving-scene image may be annotated using our “image descriptor.” As shown in Fig.~\ref{app:fig-example-annotation-truncated}, this annotation provides, for each detected object, its bounding box, its distance from the ego vehicle, and a short appearance description. Given an annotated object (e.g. ``car''), a user-specified editing action (e.g. ``replace''), and optionally a target description (e.g. ``green truck''), we form a simple sentence (i.e. ``replace the car with a green truck'') which is then encoded by CLIP. The resulting embedding is written into the mask for all pixels inside that object’s bounding box. As multiple objects may overlap, masks are assembled in order of decreasing object distance, ensuring that nearer objects overwrite farther ones.

\mypara{Interactive LangMask Construction} Alternatively, through our user interface a user may select a bounding box in the scene and specify an editing action, and optionally a target description. Then following our ``image descriptor'' a VLM would describe the selected subject appearance that we combine with the user’s inputs into a simple sentence. After encoding this instruction with CLIP, we populate the LangMask within the chosen bounding box. When multiple user selections are made, the masks are applied in order of descending object distance.

\subsection{Compound Dataset Development}\label{App:real-dataset-development}

LangMask instructions are simple sentences. Editing actions are one of \{insert, delete, modify, replace\}. For modifications, a target appearance is required (e.g. ``change the car to \ul{green}''). For replacements and insertions both a target subject and appearance are sampled (e.g. ``insert a \ul{middle-aged person}'', ``replace the car with a \ul{blue truck}'').
The LangMasks are systematically derived by comparing corresponding frame annotations
via three rules:

\begin{itemize}[leftmargin=*]
    \vspace{-0.1cm}
    \item Distance-based filtering: Objects beyond 50 meters from the ego vehicle are excluded unless they occupy a significant image area.
    \item Truncation detection for undersized 2D bounding boxes near image boundaries.
    \item Occlusion handling: In complex traffic scenarios, overlapping vehicle bounding boxes are each preserved to maintain scene coherence.
\end{itemize}

For compound editing pairs, the traffic usually completely changes between images. For this reason, we prepend in the global text prompt ``remove all pedestrians and all vehicles from the scene.'' Meanwhile the LangMask specifies the vehicles and pedestrians to insert. Traffic lights use the ``modify'' editing action.

\subsection{Pseudo-Dataset Development}\label{App:psuedo-image-pipeline}

LangMask instructions are simple sentences. Editing actions are one of \{insert, delete, modify, replace\}. For modifications, a target appearance is required (e.g. ``change the car to \ul{green}''). For replacements and insertions both a target subject and appearance are sampled (e.g. ``insert a \ul{middle-aged person}'', ``replace the car with a \ul{blue truck}'').

For pseudo-pair development, we perform one edit per LangMask. Given an unpaired image, we annotate it with our ``image descriptor'' and randomly select one subject, and sample an editing action from \{insert, delete, modify, replace\}. If required, we randomly sample a target object and appearance from the following word banks.

\textbf{Traffic Lights.} The action is always \texttt{modify}. 
The target appearance is sampled from \{\texttt{green}, \texttt{red}, \texttt{yellow}\}.

\textbf{Vehicles.} For modifications, a target appearance is sampled. For replacements and insertions both a target subject and appearance are sampled.
\begin{itemize}
    \item Target colors:
    \texttt{[red, blue, green, yellow, black, white, silver, grey]}
    \item Target objects:
    \texttt{[car, truck, bus, motorcycle with its rider, bicycle with its rider, ambulance, fire truck]}
\end{itemize}

\textbf{Pedestrians.} For insertions and replacements, we sample from both clothing types and age.  
For modifications, we change only the clothing type.

\begin{itemize}
    \item Target clothing adjectives:
    \texttt{[red, blue, green, yellow, black, white, casual, formal, businesslike, vibrant, summer, winter, sporty]}
    \item Target clothing articles:
    \texttt{[shirt, jacket, coat, sneakers, boots, hat, dress, skirt, trousers, pants, clothes]}
    \item Target ages (for modifications):
    \texttt{[young, middle-aged, elderly]}
\end{itemize}

During training, the pseudo images are used as conditioning except for object removals which with 0.5 probability are used as ground-truth to have equal representation between deletions and insertions.

\section{Training \& Inference Details}\label{app:training-details}

\subsection{Training} \label{App:training-details-ultraedit}

To incorporate the guidance from the LangMasks, we expand the VAE input channels to accept the concatenation of the input image and conditioning masks and train end-to-end. The weights of any existing convolutions are maintained and new weights are initialized as zero. The RGB-image VAE is frozen. We train at a resolution of 512 $\times$ 512 and a learning rate of $1\times 10^{-5}$.

We evaluate the UltraEdit model with the following settings:
\begin{itemize}
    \item \textbf{UltraEdit}. The UltraEdit model supports a single binary mask as conditioning therefore we project our LangMask into a binary mask. The LangMask editing prompts are appended to the global text prompt.
    \item \textbf{UltraEdit-Text}. We train an UltraEdit model using only supervised objectives without LangMasks. To do this, following the process described in Sec. ~\ref{subsec:real-world-paird-data}, we construct the editing prompts by asking chatGPT-4o to describe, in addition to global changes, all objects to remove from left to right, and all objects to add from left to right. Similar to the base model we project our LangMask into a binary mask. This model was trained only on the compound editing dataset from Sec.~\ref{subsec:real-world-paird-data}. This model is trained across 4 $\times$ 48 GB NVIDIA A6000 GPUs with a total batch size of 256 for 10000 steps. We set $\lambda_{\text{sft}} = 1.0$ during supervised training.
    \item \textbf{UltraEdit-Mask}. We train an UltraEdit model using only supervised training objectives and LangMasks. This model was trained only on the compound editing dataset from Sec.~\ref{subsec:real-world-paird-data}. We train this model across $4 \times 80$GB NVIDIA H100 GPUs with a total batch size of 4 for 25000 steps. We set $\lambda_{\text{sft}} = 1.0$ during supervised training.
    \item \textbf{UltraEdit-Mask-UFT}. We train an UltraEdit model using both supervised and unsupervised training objectives and LangMasks. This model was trained only on the compound editing dataset from Sec.~\ref{subsec:real-world-paird-data}. We finetune this model across $4 \times 80$GB NVIDIA H100 GPUs with a total batch size of 4 for 1000 steps starting at the \textit{UltraEdit-Mask} pretrained checkpoint described above. We use the same loss hyperparameters as below.
    \item \textbf{Ours}. We train an UltraEdit model using supervised training for 25000 steps on both the compound editing dataset from Sec.~\ref{subsec:real-world-paird-data} and pseudo-paired dataset (both local and global) from Sec.~\ref{sec:methods-pseudo-dataset-development}. We set $\lambda_{\text{sft}} = 1.0$ during supervised training. Then we finetune this model for 1000 steps using all the objectives described in Sec. \ref{sec:methods-training-objectives}. To adapt these objectives to multi-step diffusion, we apply gradient checkpointing and perform end-to-end training with our unsupervised losses. We train this model across $4 \times 80$GB NVIDIA H100 GPUs with a total batch size of 4 for 25000 steps. Our training parameters are:
\begin{equation}
\begin{aligned}
\lambda_{\text{gan}} &= 0.5, \\
\lambda_{\text{id}} &= 0.05, &
\lambda_{\text{id-lpips}} &= 0.05, \\
\lambda_{\text{cycle}} &= 0.05, &
\lambda_{\text{cycle-lpips}} &= 0.05, \\
\lambda_{\text{sft}} &= 3.0, &
\lambda_{\text{sft-lpips}} &= 0.5, \\
\lambda_{\text{clip}} &= 3.
\end{aligned}
\end{equation}
\end{itemize}

Supervised training requires 16 hours to complete 25,000 steps. When all the objectives described in Sec. \ref{sec:methods-training-objectives} are enabled, the model trains at approximately 15 seconds per iteration. Consequently, the 1,000-iteration post-training stage takes roughly 4 hours.

\subsection{Inference}

We perform inference on Qwen-Image-Edit, BAGEL, Omnigen2, and UltraEdit following their default hyperparameters. Generating a single image with 100 inference steps using Qwen-Image-Edit takes roughly 60 seconds on an 80GB NVIDIA H100 GPU.

For \textit{UltraEdit-Text} and \textit{UltraEdit-Mask} we follow the default hyperparameters of UltraEdit. For \textit{UltraEdit-Mask-UFT} and \textit{Ours}, we use 8 inference steps and disable classifier-free-guidance mirroring training. Generating a single image takes less than 5 seconds on an 80GB NVIDIA H100 GPU.

\section{Cycle Consistency Details}\label{app:cycle-consistency-details}

Global text prompts are either obtained by a VLM for pseudo-paired data, or by applying chatGPT-4o to our image annotations for compound data. In Section~\ref{sec:methods-training-objectives}, $t_t$ corresponds to the caption of the target image and $t_s$ corresponds to the caption of the source image. 
Each image is annotated with up to four LLM-paraphrased variations of the same caption and one is randomly sampled during training. In the case of local edits in pseudo-paired data, $t_t$ and $t_s$ match, while $M_t$ describes the transformation to obtain the target and $M_s$ describes the transformation to obtain the source.

\section{Additional Results}

\begin{figure*}[h]
  \centering
  \vspace{-2em}
  \captionsetup{font=small}
  
  \includegraphics[clip, trim=0cm 0.1cm 0cm 0cm, width=\textwidth]{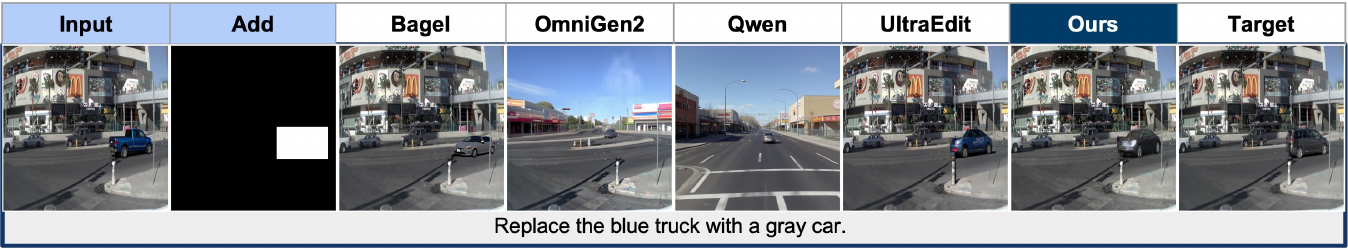}
  \includegraphics[clip, trim=0cm 0.1cm 0cm 0cm, width=\textwidth]{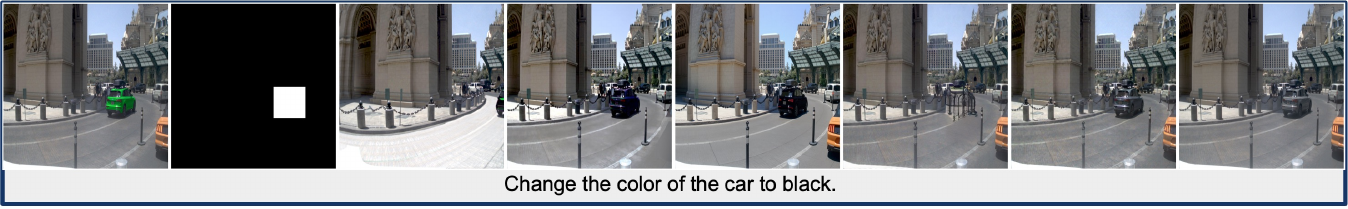}
  \includegraphics[clip, trim=0cm 0.1cm 0cm 0cm, width=\textwidth]{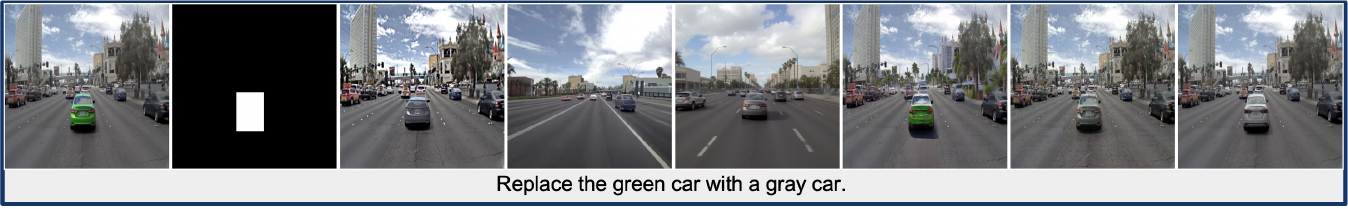}
  \includegraphics[clip, trim=0cm 0.1cm 0cm 0cm, width=\textwidth]{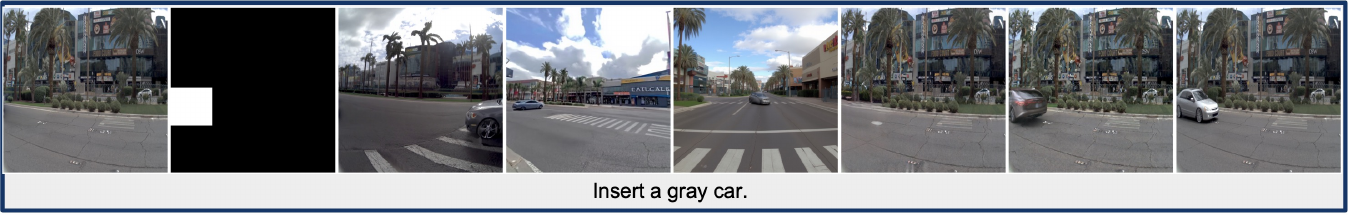}
  \includegraphics[clip, trim=0cm 0cm 0cm 0cm, width=\textwidth]{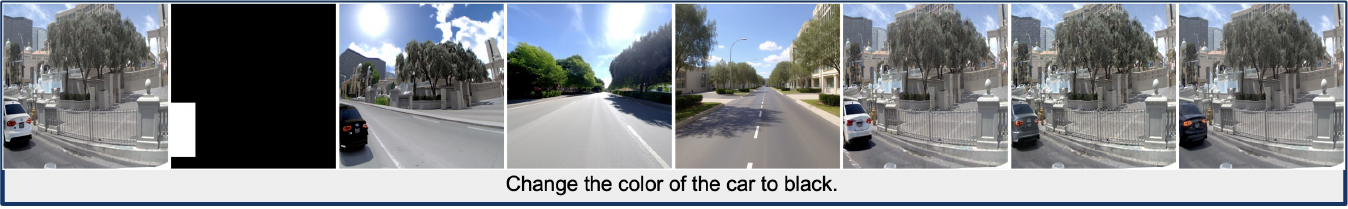}
  \includegraphics[clip, trim=0cm 0.1cm 0cm 0cm, width=\textwidth]{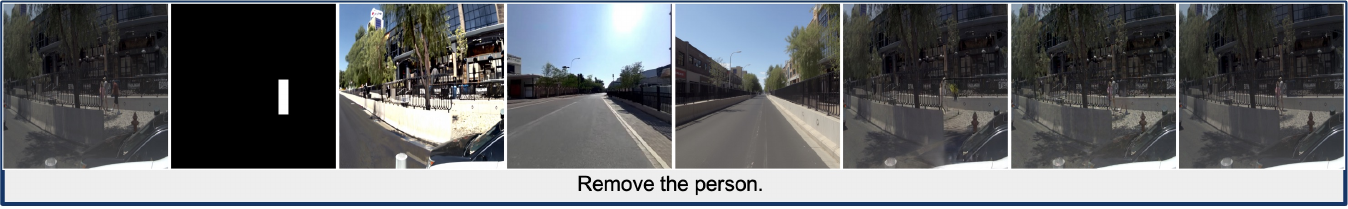}
  \includegraphics[clip, trim=0cm 0.0cm 0cm 0cm, width=\textwidth]{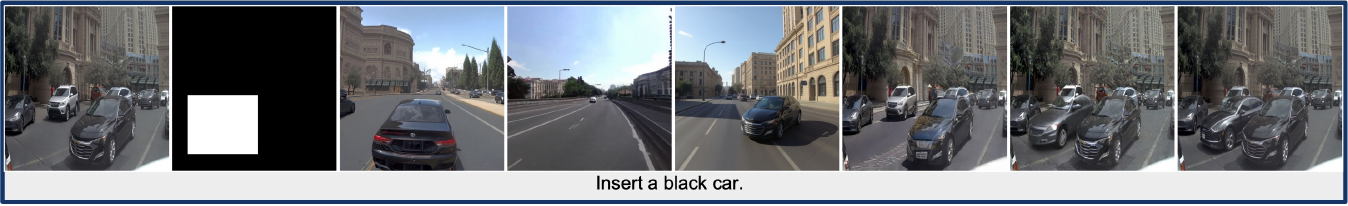}
  \captionsetup{font=small}
  \vspace{-0.5em}
  \caption{\textbf{\ourmethod Editing. Local edits:} the masks (projected as binary images and stated in text for reference) enable modifications to traffic. We compare to Qwen~\cite{wu2025qwenimagetechnicalreport}, OmniGen2~\cite{wu2025omnigen2}, UltraEdit~\cite{zhao2024ultraeditinstructionbasedfinegrainedimage}, and BAGEL~\cite{deng2025emergingpropertiesunifiedmultimodal}.}\label{app:fig-local-editing}
  \vspace{-0.5em}
\end{figure*}

\begin{figure*}[t]
  \centering
  \vspace{-2em}
  \captionsetup{font=small}
  \includegraphics[clip, trim=0cm 0.1cm 0cm 0cm, width=\textwidth]{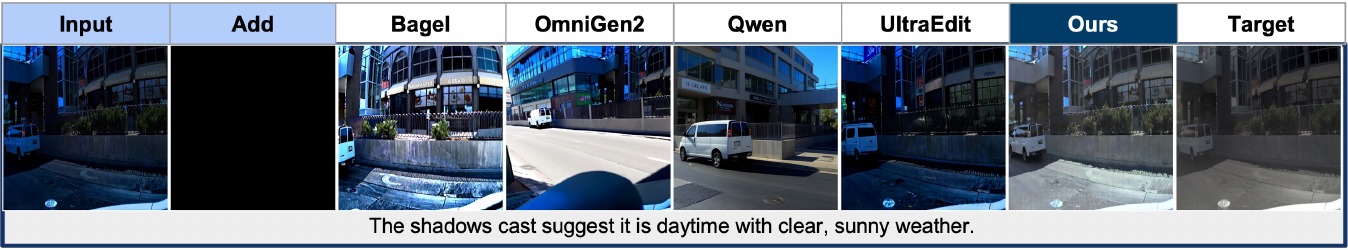}
  \includegraphics[clip, trim=0cm 0.1cm 0cm 0.8cm, width=\textwidth]{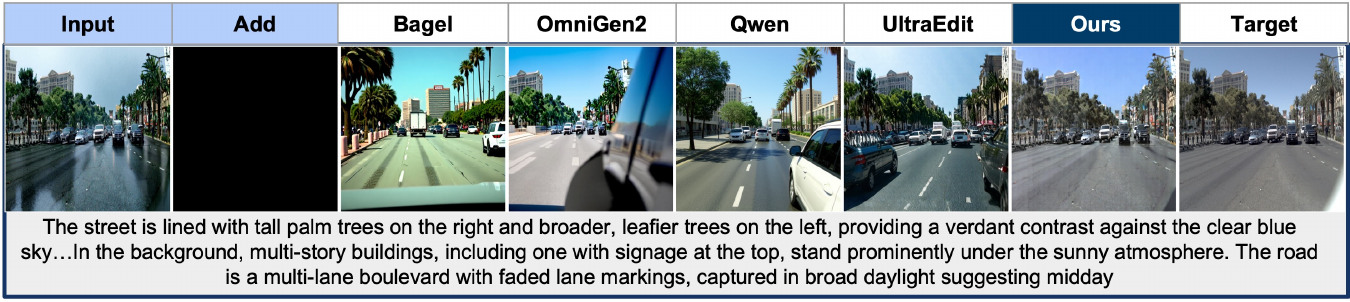}
  \includegraphics[clip, trim=0cm 0cm 0.1cm 0.8cm, width=\textwidth]{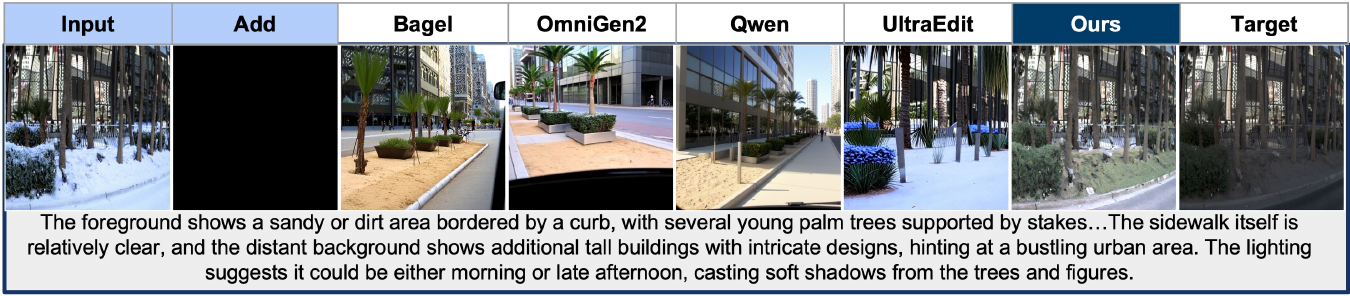}
  \captionsetup{font=small}
  \vspace{-2em}
  \caption{\textbf{\ourmethod Editing. Global edits:} the text prompt informs the appearance of the scene. For brevity, only the portions relevant to the shown edits are displayed. We compare to Qwen~\cite{wu2025qwenimagetechnicalreport}, OmniGen2~\cite{wu2025omnigen2}, UltraEdit~\cite{zhao2024ultraeditinstructionbasedfinegrainedimage}, and BAGEL~\cite{deng2025emergingpropertiesunifiedmultimodal}.}\label{app:fig-global-editing}
  \vspace{-2em}
\end{figure*}

\begin{figure*}[!b]
  \centering
  \vspace{-2em}
  \captionsetup{font=small}
    \includegraphics[clip, trim=0cm 0.1cm 0cm 0cm, width=\textwidth]{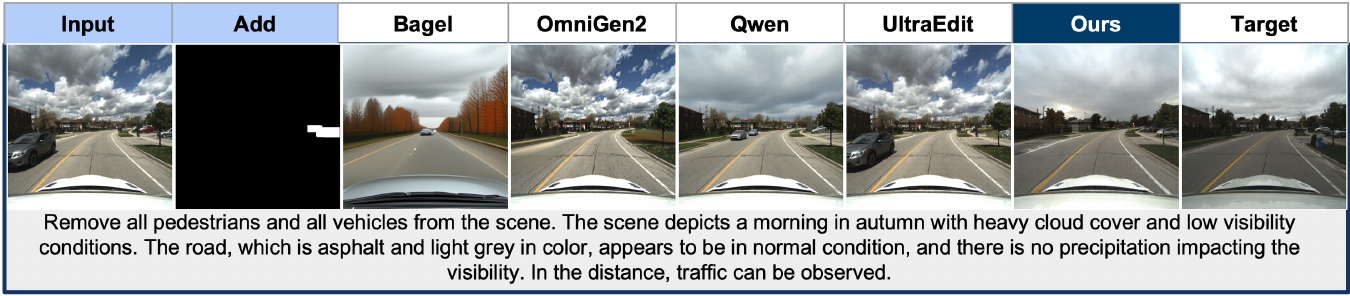}
  \includegraphics[clip, trim=0cm 0.5cm 0cm 0.8cm, width=\textwidth]{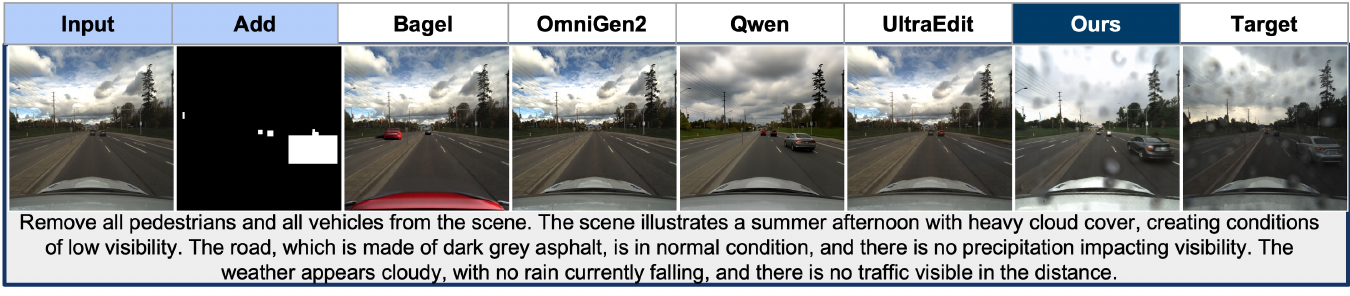}
  \includegraphics[clip, trim=0cm 0.15cm 0cm 0.8cm, width=\textwidth]{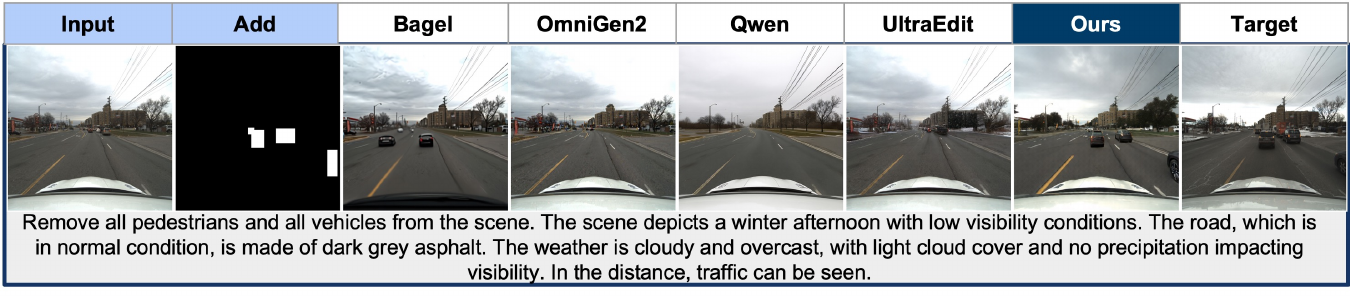}
  \includegraphics[clip, trim=0cm 0.15cm 0cm 0.8cm, width=\textwidth]{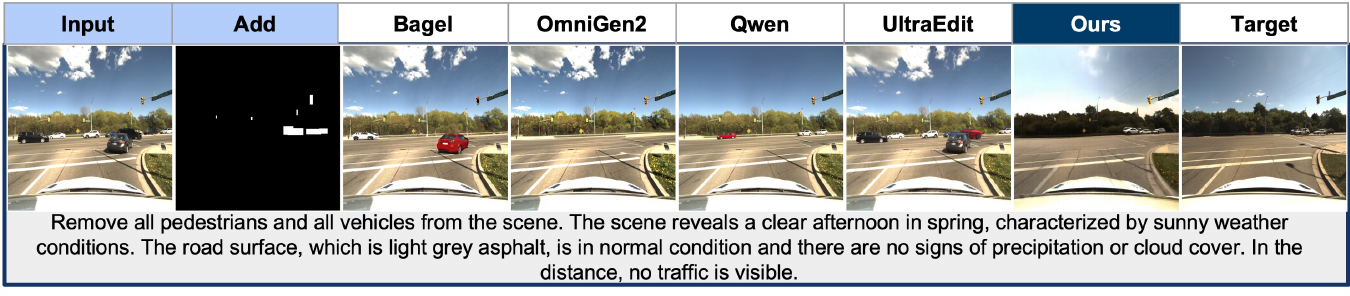}
  \captionsetup{font=small}
  \vspace{-2em}
  \caption{\textbf{\ourmethod Editing. Compound edits:} The masks (projected as binary images) enable modifications to traffic while the text prompt informs the desired global appearance. We compare to Qwen~\cite{wu2025qwenimagetechnicalreport}, OmniGen2~\cite{wu2025omnigen2}, UltraEdit~\cite{zhao2024ultraeditinstructionbasedfinegrainedimage}, and BAGEL~\cite{deng2025emergingpropertiesunifiedmultimodal}.}\label{app:fig-compound-editing}
  \vspace{-2em}
\end{figure*}

We show additional results for local editing (Fig.~\ref{app:fig-local-editing}), global editing (Fig.~\ref{app:fig-global-editing}) and compound editing (Fig.~\ref{app:fig-compound-editing}).

\section{Driving Specific Editor Baselines \& Long-tail Editing}

\begin{figure*}[t]
\vspace{-1.5em}
\centering
\includegraphics[width=0.7\textwidth]{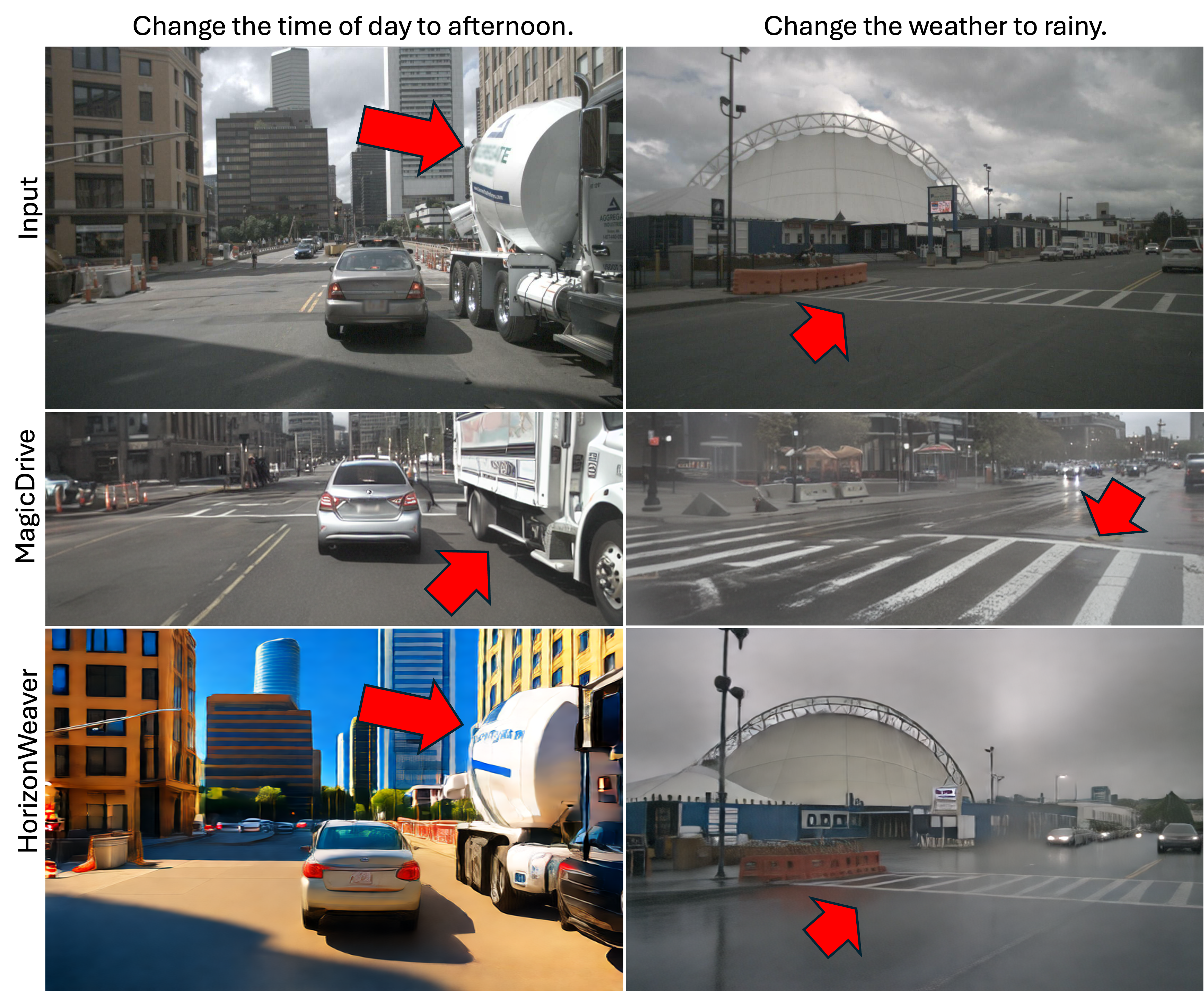}
\vspace{-1em}
\caption{\textbf{Driving specific editor.} MagicDrive \cite{gao2023magicdrive}, cannot preserve critical driving scene components (appearance of vehicles, construction infrastructure, road signs, etc.) \textbf{Long-Trail Editing} our edits can extend to rare scenarios, e.g., crosswalks, road signs, and specialized vehicles (cement mixer) which are not preserved by MagicDrive.}\label{Fig:rebuttal-magicdrive}
\end{figure*}

As shown in Fig. \ref{Fig:rebuttal-magicdrive}, MagicDrive does not take RGB images as conditions, and thus cannot preserve critical driving scene components (appearance of vehicles, construction infrastructure, road signs, etc.). Fig~\ref{Fig:rebuttal-magicdrive} demonstrates editing on rare scenarios, e.g., crosswalks, road signs, and specialized vehicles (cement mixer).

\begin{figure*}[b]
\vspace{-1em}
\centering
\includegraphics[width=0.7\textwidth]{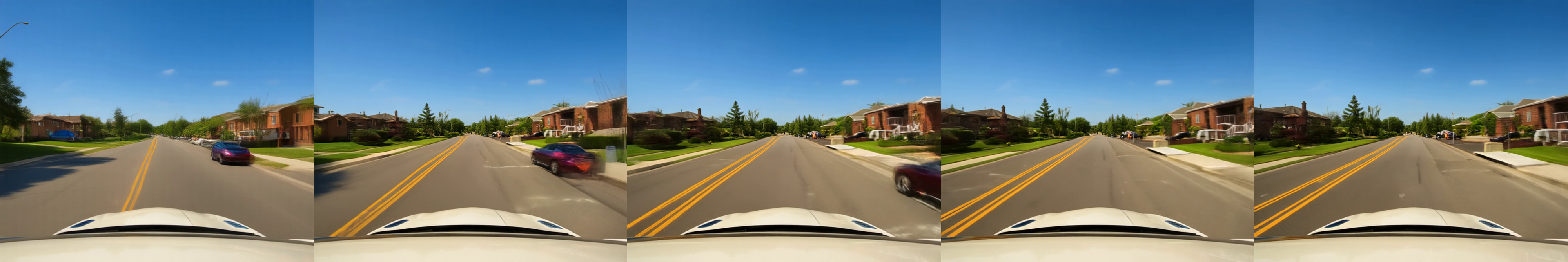}
\includegraphics[width=0.7\textwidth]{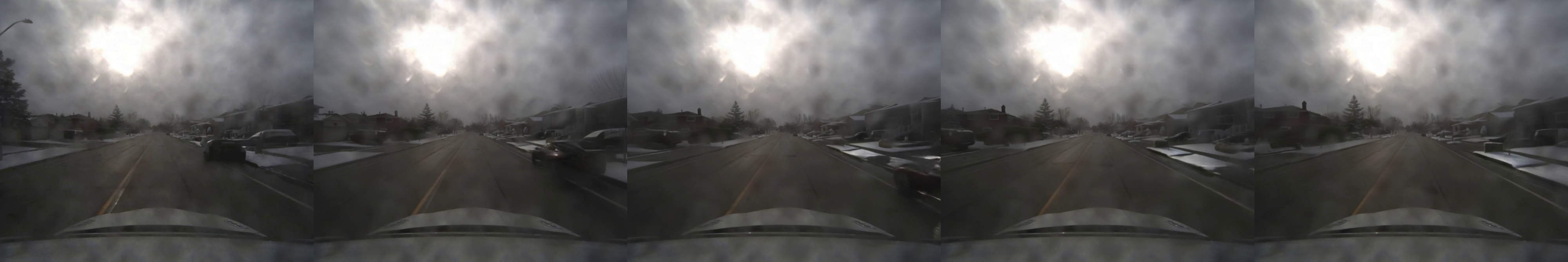}
\includegraphics[width=0.7\textwidth]{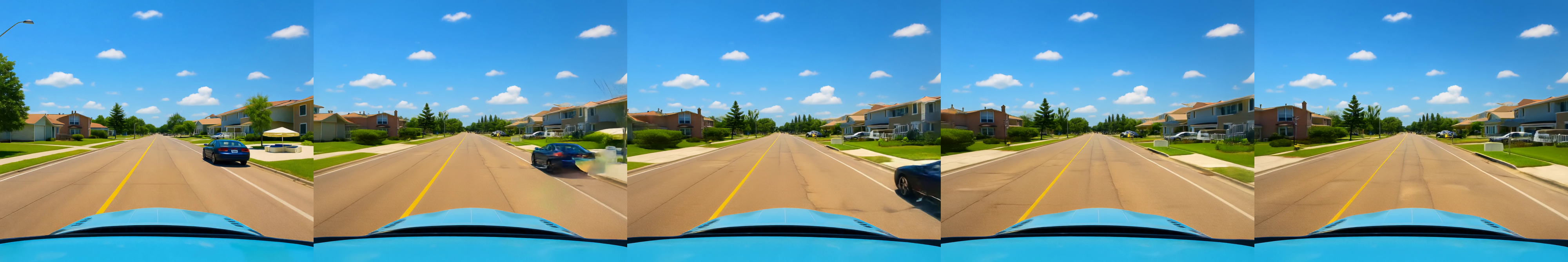}
\vspace{-0.5em}
\caption{
\textbf{Temporal Consistency.}
Videos produced from an edited initial image by \ourmethod, UltraEdit,
and Qwen (top to bottom). UltraEdit leaves the initial image unchanged.
Prompt: \textit{Change the season to summer.}
}
\label{Fig:rebuttal-temporal-consistency}
\vspace{-2em}
\end{figure*}

\section{Additional Safety-Critical Challenges in Autonomous Driving}

The application of BEV map segmentation is important and studied in multiple work (\citep{gao2023magicdrive, liu2022bevfusion, li2024bevformer}). Our results demonstrate that our synthesized weather improves BEV map segmentation in real-world conditions. 
To show that \ourmethod can help core AD challenges, such as temporal and geometric consistency, we use our edits to synthesize driving scene video edits using \cite{nvidia2025worldsimulationvideofoundation}. We show qualitative results in Fig \ref{Fig:rebuttal-temporal-consistency}.

\begin{table}[h]
\vspace{-1em}
\centering
\caption{Quantitative comparison of semantic consistency, temporal consistency, and perceptual quality.}
\vspace{-1em}
\setlength{\tabcolsep}{3pt}
\renewcommand{\arraystretch}{1.05}
\begin{tabular}{lccc}
\hline
 & UltraEdit & Qwen & Ours \\
\hline
Sem. Cons.  & 5.603 & 8.250 & 8.191 \\
Temp. Cons. & 6.588 & 6.471 & 6.765 \\
Perceptual  & 2.603 & 6.191 & 6.265 \\
\hline
\end{tabular}
\label{tab:rebuttal-metrics}
\vspace{-1em}
\end{table}

Following the evaluation in \cite{ku2024viescoreexplainablemetricsconditional} we quantify 1)
the instruction alignment
2) the temporal consistency, and
3) the perceptual quality of the video edit.
Tab. \ref{tab:rebuttal-metrics} shows that our method is competitive with Qwen in semantic consistency and outperforms baselines in temporal consistency and visual quality. Our model has fewer than 8B parameters whereas Qwen has 20B.

We further evaluate geometric consistency by comparing the depth prediction of edited images from Boreas with the ground truth LiDAR. 
To do this we run the depth-prediction model from Depth-AnythingV3 \citep{lin2025depth3recoveringvisual} and follow its evaluation metric. 
\ourmethod yields $\delta_1$ of $67.7$, which is larger by $8.9$ and $7.6$ than Qwen and UltraEdit, respectively, showing that our edits are geometry-aligned.
Many works such as \citep{gao2023magicdrive, mo2025dreamlandcontrollableworldcreation, lu2024infinicube} which target these AD data generation challenges are limited for editing as they do not enforce content preservation which includes safety-critical elements. Our editing can preserve scene critical elements Fig~\ref{Fig:rebuttal-magicdrive} such as the cement mixer and crosswalk.

\section{Computational Cost \& Scalability} We are comparable with previous models.
We train our models in two stages. Stage 1 is supervised finetuning for 25K steps (21 GPU hours), comparable to UltraEdit. Stage 2, adds our unsupervised training objectives for 2K steps (8 GPU hours) improving performance with minimal additional cost, demonstrating the scalability of our approach.
Inference is efficient: our model generates an image in 8 steps (taking 0.021s with 17GB of GPU memory), similar to UltraEdit (50 steps in 0.025s using 17GB of GPU memory), but much faster and more memory-efficient than Qwen (taking 68s using 61GB of GPU memory). Although UltraEdit has lower per-step latency, its multi-step performance is worse than ours.

\section{LLM \& VLM Evaluation} 
To analyse annotation/construction noise, we retrain our model on pre- and post-filtered data for 10K steps. Tab.~\ref{tab:rebuttal-filtering-effect} shows the performance of the editing increases with filtering, especially for fine-grained editing. 56\% of all our edited images are retained post-filter. For more analysis, please see \cite{yao2025ifinder} Sec. 4, which forms part of our LLM-based pipeline.

\begin{table}[H]
\centering
\caption{\textbf{Effect of VLM and algorithmic filtering.}}
\vspace{-1em}
\setlength{\tabcolsep}{2.5pt}
\renewcommand{\arraystretch}{1.15}
\resizebox{\columnwidth}{!}{
\begin{tabular}{lcccc|cccc}
\toprule
& \multicolumn{4}{c|}{\textbf{Fine-Grained (Full)}} 
& \multicolumn{4}{c}{\textbf{Fine-Grained (Crop)}} \\
\cmidrule(lr){2-5} \cmidrule(lr){6-9}
\textbf{Setting}
& L1↓ & L2↓ & CLIP↑ & DINO↑
& L1↓ & L2↓ & CLIP↑ & DINO↑ \\
\midrule
Prefilter
& 0.0300 & 0.0032 & 0.9423 & 0.9263
& 0.0720 & 0.0157 & 0.8214 & 0.6646 \\
Postfilter
& 0.0321 & 0.0040 & 0.9450 & 0.9267
& 0.0755 & 0.0176 & 0.8206 & 0.6536 \\
\midrule
& \multicolumn{4}{c|}{\textbf{Global}} 
& \multicolumn{4}{c}{\textbf{Compound}} \\
\cmidrule(lr){2-5} \cmidrule(lr){6-9}
Prefilter
& 0.1010 & 0.0196 & 0.9193 & 0.9112
& 0.2030 & 0.0740 & 0.8708 & 0.8247 \\
Postfilter
& 0.0983 & 0.0184 & 0.9174 & 0.9074
& 0.2026 & 0.0768 & 0.8683 & 0.8374 \\
\bottomrule
\end{tabular}
}
\vspace{-0.5em}
\label{tab:rebuttal-filtering-effect}
\end{table}

\section{LangMasks with Localized Semantics} 
\begin{figure}[H] 
  \centering
  \captionsetup{font=small}
  \includegraphics[width=0.4\textwidth]{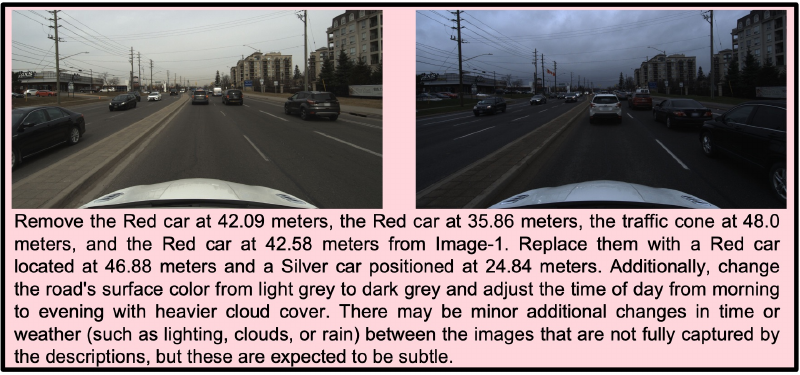}
  \captionsetup{font=small}
  \vspace{-0.4em}
  \caption{Driving scenes are dense; thus it is difficult to describe rich semantics for multiple object-oriented changes solely with natural language.}
  \label{Fig:driving-features-are-dense}
  \vspace{-1.9em}
\end{figure}

\begin{figure}[H]
  \centering
  \captionsetup{font=small}
  \includegraphics[width=0.13\textwidth]{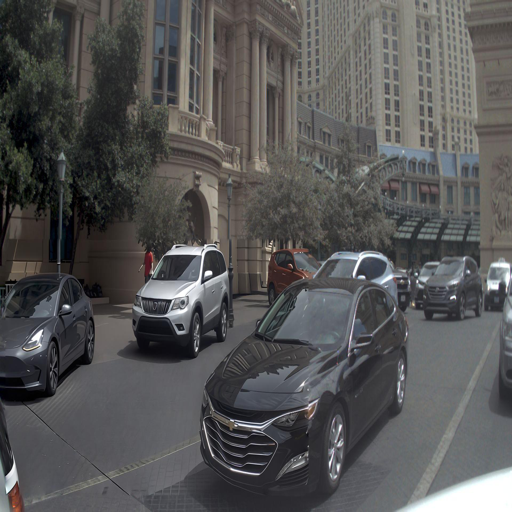}
  \includegraphics[width=0.13\textwidth]{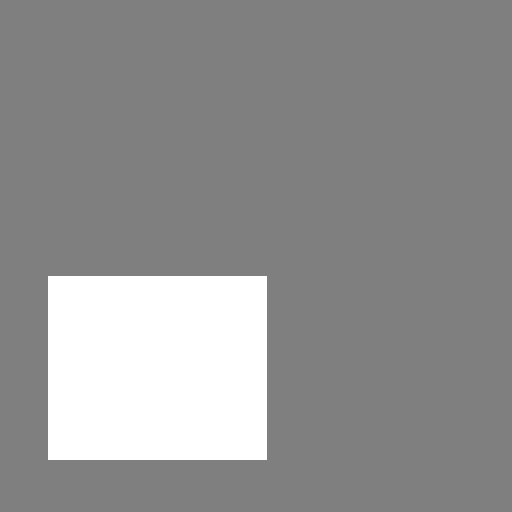}
  \includegraphics[width=0.13\textwidth]{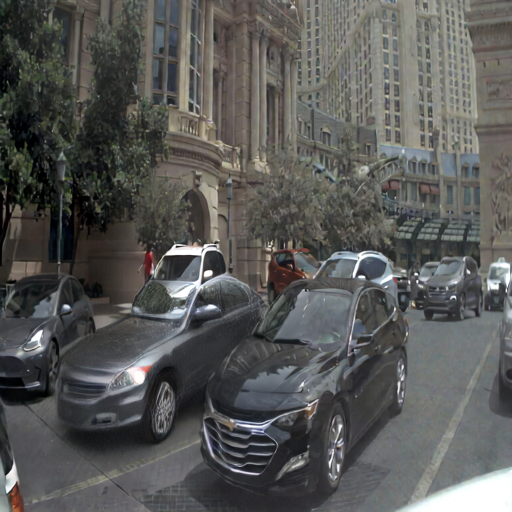}
  \vspace{-0.5em}
  \caption{\textbf{Overlapping Edits}, \ourmethod will apply insertion edits to empty spaces in crowded regions}\label{fig:rebuttal-overlapping-edits}
\end{figure}

Unlike general purpose image editing, LangMasks allow for localized semantic information to be provided to the editor. Fig. \ref{Fig:driving-features-are-dense} shows the complexity of describing dense driving scenes without LangMasks. \ourmethod can apply insertion edits to empty spaces in crowded regions as shown in Fig. \ref{fig:rebuttal-overlapping-edits}. 

\setlength{\columnsep}{4pt}

\section{Limitations}
A failure case appears in Fig. 6 (row 1, col 8), where the ground truth inaccurately labels unclear lane markings as ``well-maintained.'' However, other major changes like time-of-day are generally correct. 
While our filtering partially addresses the issue of residual artifacts in pseudo-editing (Tab.~\ref{tab:rebuttal-filtering-effect}), we plan to further improve it in future work.






\end{document}